\documentclass[10pt,twocolumn,letterpaper]{article}

\usepackage{wacv}
\usepackage{times}
\usepackage{epsfig}
\usepackage{graphicx}
\usepackage{amsmath}
\usepackage{amssymb}
\usepackage{xcolor}
\usepackage{caption}
\usepackage{enumitem}
\usepackage{multirow}
\usepackage[linesnumbered,ruled,vlined]{algorithm2e}
\usepackage{booktabs}
\usepackage{paralist}

\SetKwInput{KwInput}{Input}                
\SetKwInput{KwOutput}{Output} 

\definecolor{Green}{rgb}{0.0, 0.5, 0.0}
\definecolor{Red}{rgb}{1, 0.6, 0.6}
\definecolor{Blue1}{rgb}{0, 0.6, 1}
\definecolor{Red1}{rgb}{0.5, 0.2, 0.2}
\definecolor{PaleYellow}{rgb}{0.8, 0.8, 0}

\long\def\ignorethis#1{}

\newcommand{\Paragraph}[1]{\vspace{1mm}\noindent\textbf{#1}}

\wacvfinalcopy 

\usepackage[breaklinks=true,bookmarks=false]{hyperref}

\begin{document}

\title{Real-time Localized Photorealistic Video Style Transfer}

\author{Xide Xia*\\
Boston University\\
{\tt\small xidexia@bu.edu}
\and
Tianfan Xue\\
Google Research\\
{\tt\small tianfan@google.com}
\and
Wei-sheng Lai\\
Google\\
{\tt\small wslai@google.com}
\and
Zheng Sun\\
Google Research\\
{\tt\small zhengs@google.com}
\and
Abby Chang\\
Google\\
{\tt\small abbychang@google.com}
\and
Brian Kulis\\
Boston University\\
{\tt\small bkulis@bu.edu}
\and
Jiawen Chen\\
Google Research\\
{\tt\small jiawen@google.com}
}

\makeatletter
\g@addto@macro\@maketitle{
    \vspace*{-21pt}
    \centering
    \captionsetup{type=figure}
    \includegraphics[width=\linewidth]{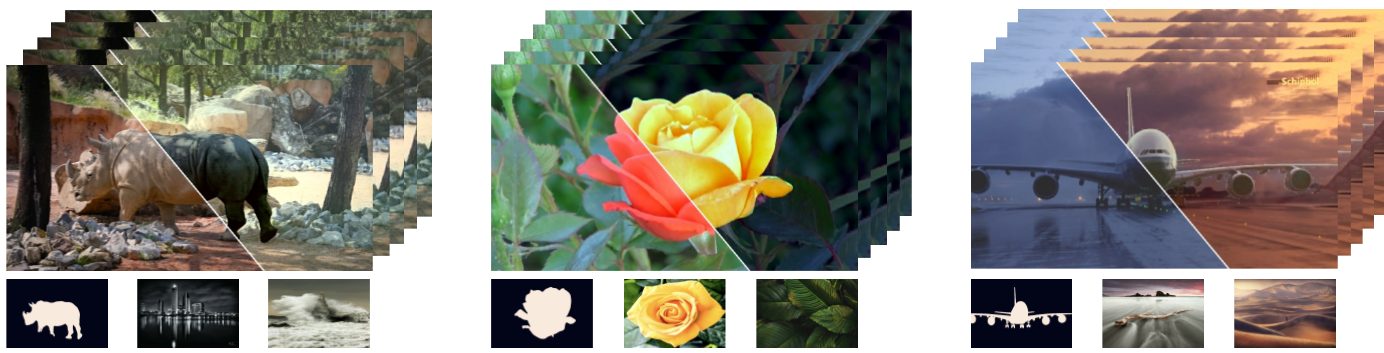}
    \captionof{figure}{\textbf{Photorealistic video style transfer}. Given an input content video and its object segmentation mask, our method learns to transfer different styles to different local regions while preserving the photorealism and temporal consistency (left: original, right: stylized). The bottom of each example shows the object mask, foreground style image, and background style image, respectively.}
	\label{fig:teaser} \vspace{-0.0cm}
    \vspace*{4pt}
}
\makeatother
\maketitle

\begin{abstract}
We present a novel algorithm for transferring artistic styles of semantically meaningful local regions of an image onto local regions of a target video while preserving its photorealism. Local regions may be selected either fully automatically from an image, through using video segmentation algorithms, or from casual user guidance such as scribbles. Our method, based on a deep neural network architecture inspired by recent work in photorealistic style transfer, is real-time and works on arbitrary inputs without runtime optimization once trained on a diverse dataset of artistic styles. By augmenting our video dataset with noisy semantic labels and jointly optimizing over style, content, mask, and temporal losses, our method can cope with a variety of imperfections in the input and produce temporally coherent videos without visual artifacts. We demonstrate our method on a variety of style images and target videos, including the ability to transfer different styles onto multiple objects simultaneously, and smoothly transition between styles in time.\let\thefootnote\relax\footnotetext{* Work done while interning at Google Research.}
\end{abstract}


\section{Introduction}
\label{sec:intro}
Color stylization plays a critical role in modern cinematography and video storytelling. It has the powerful ability to grab audience attention, elicit emotions, and convey implicit mood. For example, red usually evokes ideas of action, adventure, and strength; orange can represent joy, creativity and stimulation; and green signifies envy and health.
%
In additional to applying color styles globally to the entire video, modern filmmakers often utilize localized color tone changes; i.e., distinct color palettes applied to certain segmented objects in the scene, as a powerful tool in video storytelling. Some well-known examples include the film \textit{Schindler's List}~\cite{schindlerList}, in which the famous scene of a girl in a red coat becomes the most memorable symbol of the film. Similarly, the film \textit{Sin City}~\cite{sincity} uses high saturation (such as in red and yellow) to colorize certain characters or clothes, contrasting with the film's classic black-and-white noir theme.

While color stylization is critical to video production, color style creation and editing today is still a labor-intensive and time-consuming process even with the use of professional editing software. Many operations are still manual in the editing process, such as rotoscoping, refinement of segmentation masks, ensuring temporal consistency from frame to frame, and carefully fine-tuning color tones. An automatic approach would be incredibly helpful.

To automatically apply color styles to videos, a number of methods have recently been introduced based on advances in deep learning. For instance, several methods have explored the transfer of artistic styles to images or videos ~\cite{gatys2016image, johnson2016perceptual,li2017universal, huang2017arbitrary,sheng2018avatar, chen2020optical, Texler20_SIG}. However, due to their nature as \textit{artistic} style transfer methods, they all introduce undesirable painterly spatial distortions. 
Another line of work \cite{luan2017deep,li2018closed,li2019learning,yoo2019photorealistic, xia2020joint} focuses on \textit{photorealistic} style transfer, which requires the output to maintain ``photorealism''; i.e., the output should look as if it was taken by a real camera (like most stylized films).
However, since they primarily target still photography, existing methods often generate visible flicker artifacts when they are applied to videos.
To reduce temporal instability, several methods have tried minimizing an optical flow warping loss~\cite{huang2017real,gupta2017characterizing} or sequentially propagating intermediate features~\cite{lai2018learning}.
To our knowledge, there is no existing work that can successfully perform photorealistic localized style transfer on videos with reasonably good runtime speed.

In this paper, we present a novel approach that simultaneously addresses three major challenges in localized photorealistic video style transfer: 1) spatial and temporal style coherence over time, 2) robustness against imperfect segmentation masks, and 3) high-speed processing.
To achieve high-speed performance, we extend our approach from the recent work of Xia~et~al.~\cite{xia2020joint}, which learns local edge-aware affine transforms from low-resolution content and style inputs, with the results sliced out from a compact transform representation at the full resolution.
To minimize spatial artifacts and improve the temporal coherence, we propose a novel spatiotemporal feature transfer layer (ST-AdaIN) that is able to transfer style to local regions and generate temporally coherent stylized videos.
To handle imperfect segmentation masks, our algorithm learns an enhancement network to improve the boundaries of masks. Moreover, we compose foreground and background color transform coefficients in the grid space, resulting in high quality results even if segmentation masks are inaccurate.
Experimental results demonstrate that our model generates stylized videos with fewer visual artifacts and better temporal coherence than existing photorealistic style transfer methods.

In summary, our contributions in this work are threefold:
\begin{compactitem}
\item A differentiable spatiotemporal style transfer layer (ST-AdaIN) to match the feature statistics of local regions, generating temporally coherent stylized results (Section~\ref{subsec:spatiotemporal}).
\item Mask enhancement and grid-space blending algorithms that can render natural style transitions between objects given noisy selection masks (Section~\ref{subsec:mask_enhancement}).
\item A deep neural network for photorealistic video style transfer that runs in real-time (26.5~Hz at $1024\times1024$ resolution).
%
\end{compactitem}



\section{Related Work}

\Paragraph{Image style transfer.}
Classical style transfer approaches can be categorized into: 1) global methods, which match global image statistics~\cite{reinhard2001color, pitie2005n}, and 2) local methods~\cite{laffont2014transient, shih2014style, shih2013data, wu2013content, tsai2016sky}, which find dense correspondence between content and style.
While the global methods are efficient, the results are not always faithful to the style image.
The local methods can generate high-quality results, but they are computationally expensive and often limited to specific scenarios (e.g., portraits, sky, or season changes).

Convolutional neural network (CNN)-based style transfer has been widely studied in recent years.
The pioneering work of Gatys et al.~\cite{gatys2016image} formulates the problem as an iterative optimization to match the statistics of feature activations within a pre-trained classification network.
Several follow-up work either improves stylization efficiency by learning strictly feed-forward networks~\cite{johnson2016perceptual,li2016precomputed}, or increase generalization by adding the ability to transfer multiple styles within a single ``universal'' model~\cite{huang2017arbitrary,sheng2018avatar,li2017universal}.
Although these approaches produce impressive artistic stylization results, the images often contain
spatial distortions and warped image structures that are unacceptable in a photorealistic setting.

To achieve photorealistic style transfer, Luan et al.~\cite{luan2017deep} extend the optimization framework of Gatys~et~al.~\cite{gatys2016image} by imposing a local affine \emph{photorealism constraint}.
PhotoWCT~\cite{li2018closed} optimizes the loss functions of Luan et al.~\cite{luan2017deep} in a closed-form solution but requires a post-processing step to further smooth the stylized results.
Li~et~al.~\cite{li2019learning} learn linear transformations but require a spatial propagation network~\cite{liu2017learning} as an anti-distortion filter.
On the other hand, WCT\textsuperscript{2}~\cite{yoo2019photorealistic} adopts the wavelet corrected transfer based WCT to preserve the photorealism without any post-processing.
An~et~al.~\cite{an2019ultrafast} adopt neural architecture search and network pruning to learn a lightweight model, but their model is still behind real-time performance.
Recently, Xia et al.~\cite{xia2020joint} propose a model that strictly enforces Luan's photorealism constraint by directly learning style transfer in a bilateral space~\cite{gharbi2017deep}, which can achieve real-time performance at 4K resolution.
The proposed model builds upon this fast method but is able to transfer styles to local regions and generate temporally coherent video results.

\Paragraph{Video style transfer.}
As style transfer methods typically generate new texture or change image color significantly, applying existing algorithms to videos frame-by-frame often lead to temporally inconsistent results.
%
To improve the temporal stability, several methods minimize a temporal loss (i.e., optical flow warping error~\cite{bonneel2015blind}) in an iterative optimization framework~\cite{ruder2016artistic}, training feed-forward networks~\cite{huang2017real, gupta2017characterizing}, or learning a post-processing module~\cite{lai2018learning}.
Chen et al.~\cite{chen2017coherent} propagate long-range information by blending the intermediate features with the ones from the previous frame.
Our method enforces short-temporal consistency by minimizing the temporal loss and imposes long-term temporal consistency by propagating the intermediate features of the neighboring frames through the proposed spatiotemporal feature transfer layer. 

\Paragraph{Bilateral grid.}
Bilateral space is commonly used for fast image processing. Paris and Durand~\cite{paris2006fastapprox} first introduce bilateral space for fast edge-aware image filtering, showing that it can be sampled at a low resolution \emph{bilateral grid}. Chen~et~al.~\cite{chen2016bilateral} extend the bilateral grid for fast approximation of many edge-aware image transformations.
%
Recently, Gharbi et al.~\cite{gharbi2017deep} learn a deep network to predict bilateral grid for approximating several image transformations.
The grid is only estimated from a low-resolution input and then applied to the full-resolution input image, significantly accelerating the speed of complicated image transformation.
Xia et al.~\cite{xia2020joint} further extend this structure to photorealistic image style transfer by integrating with the Adaptive Instance Normalization (AdaIN)~\cite{huang2017arbitrary}. 
In this work, we also propose to model the style transformation using bilateral grids,
and we will describe the technical differences and improvement against Xia et al~\cite{xia2020joint} in Section~\ref{sec:method}.

\begin{figure*}
\centering
\includegraphics[width=0.9\linewidth]{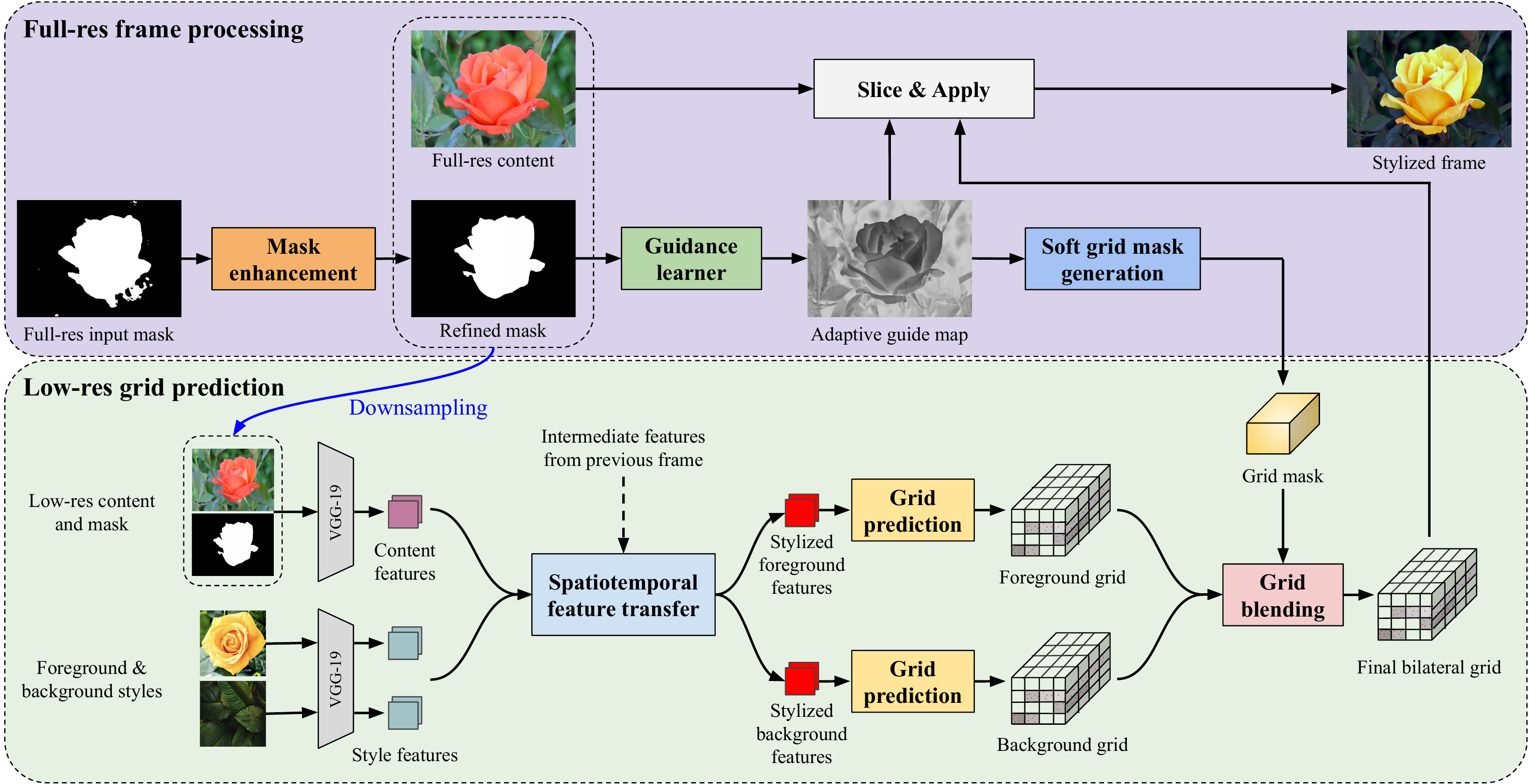}
\vspace{-2mm}
\caption{\textbf{Model architecture.}
Our model consists of two paths. The \emph{grid prediction} path consumes low-resolution content, style, and masks and infers a bilateral grid that encodes how to transform each layer. At full resolution, we predict a guide map that tells us how to best blend the style transforms, as well as how to index into it for rendering. See Section~\ref{sec:method} for details.
}
\label{fig:architecture}
\end{figure*}

\section{Method}
\label{sec:method}

We aim to transfer the style from a set of images to different local regions in a target video, producing a smooth, temporally-consistent, and photorealistic result.
We take as input a video (the \emph{content}), object selection masks (one video per object), and a set of style images (each style corresponds to one object), and output a video sequence.
Since manual rotoscoping to produce selection masks is tedious, and automatic object detection is rarely pixel-perfect, one critical design goal in localized video style transfer is the ability to handle imperfect and noisy masks.
%
Without loss of generality, we discuss the case of having two regions that we call foreground (i.e., where the mask value is $1$) and background (i.e., the complement).
Extending the algorithm to three or more regions is straightforward with linear complexity.

\subsection{Overview}
Our method builds upon the style transfer algorithm of Xia~et~al.~\cite{xia2020joint}, which we summarize here.
The algorithm consists of two ``paths'' (Figure~\ref{fig:architecture}), a full-resolution frame processing path (in purple) and a low-resolution grid prediction path (in green).
The grid prediction path takes as input a low-resolution content/style pair (e.g., $256 \times 256$), extracts features using a pre-trained VGG-19 network~\cite{Simonyan15}, and uses a series of Adaptive Instance Normalization (AdaIN) layers~\cite{huang2017arbitrary} to predict a photorealistic style transformation operator.
The transformation operator is encoded as a very low-resolution affine bilateral grid $\Gamma$ (e.g., $16 \times 16 \times 8$).
To render a full-resolution frame in real-time, the full-resolution path uses a per-pixel learned lookup table to first predict a \emph{guide map}.
The guide map serve as a learned proxy for image luminance that better separates edges.
To render a given input pixel at position $(x, y)$ and color $(r, g, b)$, the algorithm computes its guide value $z = LUT(r,g,b)$, \emph{slices} the affine bilateral grid $\Gamma$ at $(x, y, z)$ (sampling with trilinear interpolation and scaling each axis) to retrieve a $3 \times 4$ affine transform $A$, and computes the product $A \cdot (r, g, b, 1)^T$.
%
%
%
Our work is motivated by the fact that while Xia~et~al.'s algorithm is fast and works well on still photos, our early experiments showed that it is nontrivial to extend the method to localized video style transfer.
A naive localized extension would be to simply run the algorithm twice on each frame: once per region, and then blend using object masks.
However, this does not work well due to the following reasons:
\begin{compactitem}
    \item Content statistics upon which stylization operates need to be localized and temporally coherent to produce a satisfactory output. This requires a new feature transfer operator (Section~\ref{subsec:spatiotemporal}).
    \item Pixel-space blending using object masks depends critically on mask quality, which is rarely satisfactory. We address this by learning to enhance masks (Section~\ref{subsec:mask_enhancement}), which are then incorporated into the rendering process (Section~\ref{sec:stylized_rendering}).
    \item Although Xia et al.~\cite{xia2020joint} show a few smooth video results despite being trained on only still imagery, we find that without additional temporal regularization, there is noticeable flicker. We address this with the proposed spatiotemporal feature transfer layer and a warping loss (Section~\ref{subsec:spatiotemporal} and~\ref{subsec:training_losses}).
\end{compactitem}

\subsection{Spatiotemporal Feature Transfer}
\label{subsec:spatiotemporal}
Recall that AdaIN~\cite{huang2017arbitrary} transfers the mean $\mu(\cdot)$ and variance $\sigma(\cdot)$ feature statistics from $y$ (e.g., style) onto $x$ (e.g., content) as:
\begin{equation}
    \label{eq:adain}
    \mathrm{AdaIN}(x, y) = \sigma(y)\left(\frac{x-\mu(x)}{\sigma(x)}\right) + \mu(y).
\end{equation}
Mean and variance are computed independently for each feature channel.
%
Note that the original AdaIN operator computes mean and variance over an entire frame. This is inappropriate for localized style transfer as it mixes the features of foreground and background regions, leading to visible halo artifacts near object boundaries (see Figure~\ref{fig:pixel_stitching}(d) for an example).
We address this with a \textit{spatially-aware AdaIN} (SA-AdaIN), which extends the operator to account for the content mask $m_x$:
\begin{align}
    \label{eq:sa-adain}
    &\text{SA-AdaIN}(x, y, m_x)\!=\!\sigma(y)\!\left(\!\frac{x - \mu(x \cdot m_x)}{\sigma(x \cdot m_x)}\!\right)\!+\!\mu(y),
\end{align}
Next, we tackle motion. Although using AdaIN on VGG-19 features has shown to be effective for still photo style transfer~\cite{huang2017arbitrary,xia2020joint}, it has been shown that these features are sensitive to small changes in the input. Subtle motions can lead to large differences in the features~\cite{gupta2017characterizing} which are amplified by downstream stylization that appears as flicker artifacts when viewed at framerate~\cite{lai2018learning}.
To address this, we propose a \textit{temporally coherent AdaIN} (TC-AdaIN) layer that computes the mean and variance from both the previous and current frames:
\begin{equation}
    \label{eq:tc-adain}
    \text{TC-AdaIN}(x^{t}, x^{t-1}, y, \alpha) = \sigma(y)\left(\frac{x-\mu^{t}}{\sigma^{t}}\right) + \mu(y),
\end{equation}
where $\mu^{(t)} = (1-\alpha)\mu(x^{(t)})+\alpha\mu(x^{(t-1)})$, $\sigma^{(t)} = (1-\alpha)\sigma(x^{(t)})+\alpha\sigma(x^{(t-1)})$, and $\alpha$ is a weight that balances the contribution of the previous ($t - 1$) and current ($t$) frame.

Combining~\eqref{eq:sa-adain} and~\eqref{eq:tc-adain}, we obtain a \textit{spatiotemporal AdaIN} (ST-AdaIN) operator:
\begin{align}
    \label{eq:st-adain}
    &\text{ST-AdaIN}(x^{t}, x^{t-1}, y, m_x^{t}, m_x^{t-1}) = \sigma(y) \left( \frac{x - \hat{\mu}^{t}}{\hat{\sigma}^{t}} \right) + \mu(y),
\end{align}
where 
\begin{align}
    \hat{\mu}^{t} &= (1 - \alpha) \mu(x^{t} \cdot m_x^{t}) + \alpha \mu(x^{t-1} \cdot m_x^{t-1}), \\
    \hat{\sigma}^{t} &= (1 - \alpha) \sigma(x^{t} \cdot m_x^{t}) + \alpha \sigma(x^{t-1} \cdot m_x^{t-1}).
\end{align}
%
%
After the feature transferring, we apply a few convolutional layers to predict the bilateral grids for both the foreground and background.
The detailed architectures of our spatiotemporal feature transfer and grid prediction are provided in the supplementary material.

\subsection{Mask Enhancement}
\label{subsec:mask_enhancement}
In practice, input selection masks are rarely pixel-perfect, either due to errors made by object segmentation algorithms (e.g.,~\cite{long2015fully,chen2017deeplab,deeplabv3}) or inaccuracies in users' annotations.
To ensure that masking errors are not propagated to our stylized output, we introduce a mask enhancement network that smooths and aligns masks to object boundaries.
Our mask enhancement network is a lightweight module, consisting of three convolutional layers, where the last layer adopts a sigmoid activation to produce a soft boundary.
%
%

\begin{figure*}
\centering
\includegraphics[width=0.9\linewidth]{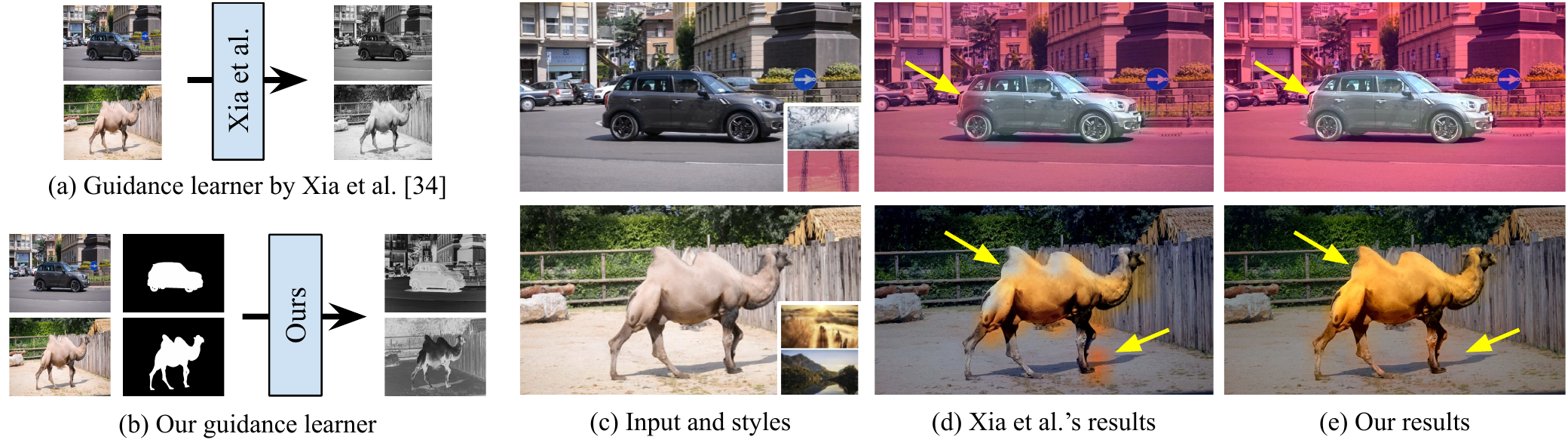}
\caption{\textbf{Guide map for grid blending and slicing.} While Xia et al.\cite{xia2020joint} learns the guidance from the input image alone, our model takes the object mask as an extra input, which allows it to better separate foreground and background at boundaries. 
}
\label{fig:new_guide}
\end{figure*}

\subsection{Stylized Rendering}
\label{sec:stylized_rendering}

After learning bilateral grids that encode foreground and background styles, we need to smoothly blend between them to produce a photorealistic output. A naive solution would be to apply the grids independently to the input frame to produce two stylized outputs, and alpha blend them using the enhanced mask calculated in Section~\ref{subsec:mask_enhancement}. 
However, applying the transformation twice at full-resolution is not only inefficient, but if the two styles (and thus transformations) are dramatically different, it will also lead to undesirable halo artifacts (see Figure~\ref{fig:new_guide}(d)) despite blending using a soft mask.
We devise a better solution that interchanges the steps: first \emph{blend the transforms} taking masks into account, then apply the blended transform once at full resolution.

Recall that HDRnet~\cite{gharbi2017deep} and bilateral photo style transfer~\cite{xia2020joint} learn a global lookup table based on pixel color alone to produce its guide map.
The guide map, which is directly used as the $z$ index when slicing the bilateral grid, serves to separate object edges.
This is insufficient in our application since object edges are delineated not only by color gradients of large magnitude, but also by explicit masks.
Therefore, we replace the learned lookup table with a 2-layer convolutional neural network that also incorporates the object mask to learn an adaptive guide map that better separates foreground and background, as shown in Figure~\ref{fig:new_guide}(a).

We use the adaptive guide map to build a soft \emph{grid-space} mask (see the supplementary material for details) and combine the two stylization transforms into a single compact representation as:
\begin{equation}
    \label{eq:soft_blending}
    \Gamma =  {M}_{grid} \cdot \Gamma_f + (1-{M}_{grid}) \cdot \Gamma_b,
\end{equation}
where $\Gamma_f $ and $\Gamma_g$ are the foreground and background grids, respectively.
Finally, we render the output at each pixel by slicing out an affine transform from blended grid $\Gamma$ using the adaptive guide map and performing a matrix multiply.
Figure~\ref{fig:new_guide}(c) and (d) shows how grid-space blending eliminates artifacts from the naive approach.

\subsection{Training Losses}
\label{subsec:training_losses}
Given a collection of input content videos $\{I_c\}$ with object masks, foreground and background style images $F_s$ and $B_s$, our model can be trained to generate a stylized output video $\{O\}$ on given an arbitrary input by optimizing the following multi-objective loss function:
\begin{equation}
\label{eq:loss_total}
\mathcal{L} = \lambda_c \mathcal{L}_c + \lambda_{s} \mathcal{L}_{s} + \lambda_r \mathcal{L}_r  + \lambda_m \mathcal{L}_{m} + \lambda_z \mathcal{L}_{z}  + \lambda_t \mathcal{L}_{t}.
\end{equation} 
We empirically set $\lambda_c=0.2$, $\lambda_s=1$, $\lambda_r=0.02$, $\lambda_m=5.0$, $\lambda_z=1.5$, and $\lambda_t=1000$ in all experiments. We discuss each component below.

We adopt the now standard \emph{content loss} that minimizes the difference between the intermediate features from a pre-trained VGG-19 network~\cite{gatys2016image}:
\begin{align}
    \mathcal{L}_{c} = \sum_{i=1}^{N_C} \left\| \Phi_{i}[O] - \Phi_{i}[I_{c}] \right\|_{F}^{2},
\end{align}
where $\Phi_i[\cdot]$ is a feature map, $N_C$ denotes the number of layers selected from VGG-19 to build content loss, and $\|\cdot\|_F$ is the Frobenius norm. Our \emph{style loss} uses the simpler formulation from recent work~\cite{huang2017arbitrary,xia2020joint} that uses only the mean and variance of of VGG-19 features:
\begin{align}
\label{eq:style_loss}
\mathcal{L}_{s} &= \sum_{i=1}^{N_S} \left\|\mu(\Phi_i[O]) - \mu(\Phi_i[I_s])\right\|_{2}^{2} \nonumber \\
&+ \sum_{i=1}^{N_S} \left\|\sigma(\Phi_i[O]) - \sigma(\Phi_i[I_s])\right\|_{2}^{2},
\end{align}
where $N_s$ denotes the number of layers selected from VGG-19 to build style loss.

To penalize abrupt local changes in the predicted transforms but allow changes across strong object edges (known as \emph{bilateral smoothness}), we adopt the \emph{Laplacian regularizer} of~\cite{xia2020joint} on both the foreground grid $\Gamma_f$ and background grids $\Gamma_b$:
\begin{align}
\label{eq:laplacian_reg}
\mathcal{L}_r &= \sum_s \sum_{t \in N(s)} ||\Gamma_f[s] - \Gamma_f[t]||_F^2 \nonumber \\
&+ \sum_s \sum_{t \in N(s)} ||\Gamma_b[s] - \Gamma_b[t]||_F^2.
\end{align}
where $N(s)$ denotes the neighbors of cell $s$.


\Paragraph{Mask loss.}
To ensure that the network does not predict an affine transformation that applies the foreground style to a background pixel and vice versa, we add a \emph{mask loss} that says: if we sliced out a full-resolution pixel-space mask from the predicted grid mask using our learned guide map, it should align with object boundaries. Mathematically:
\begin{equation}
\label{eq:mask_loss}
\mathcal{L}_{m} = \left\| slice(z, M_{grid}) - M_{gt} \right\|_{2}^{2},
\end{equation}
where $z$ is the learned guide map, $M_{grid}$ is the predicted soft grid mask, and $M_{gt}$ is the ground-truth object mask. 

\Paragraph{Guide loss.} We penalize guide maps that are substantially different from input luminance $I_{gray}$ since luminance edges are a strong segmentation signal:
\begin{equation}
\label{eq:guide_loss}
\mathcal{L}_z =  \left\| z - I_{gray} \right\|_{2}^{2}.
\end{equation}

\Paragraph{Temporal loss.}
To improve temporal coherence, we minimize the flow warping error between the output frames~\cite{lai2018learning}:
\begin{equation}
\label{eq:st_loss}
\mathcal{L}_{t} = \sum_{t=1}^{T} \sum_{i=1}^N V_{t\rightarrow t-1}^{i}||O_t^{i}-\hat{O}_{t-1}^{i}||_1,
\end{equation}
where $\hat{O}_{t-1}$ is frame $O_{t-1}$ warped by optical flow $f_{t\rightarrow t-1}$ and $V_{t\rightarrow t-1}$ is the visibility mask calculated from the input frames $I_t$ and $I_{t-1}$.
We use the PWC-Net~\cite{Sun2018PWC} to compute optical flow on the content video. 
More implementation details are provided in the supplementary material.



\section{Experimental Results}
\label{sect:exp}

In this section, we first provide qualitative comparisons with state-of-the-art photorealistic style transfer methods.
We then present quantitative results from our performance benchmark and user study.
Finally, we analyze the contribution of different components in our model.

\begin{figure*}
\centering
\footnotesize
\begin{tabular}{c@{\hspace{0.005\linewidth}}c@{\hspace{0.005\linewidth}}c@{\hspace{0.005\linewidth}}c}
  \includegraphics[width = .33\linewidth,height=0.17\linewidth]{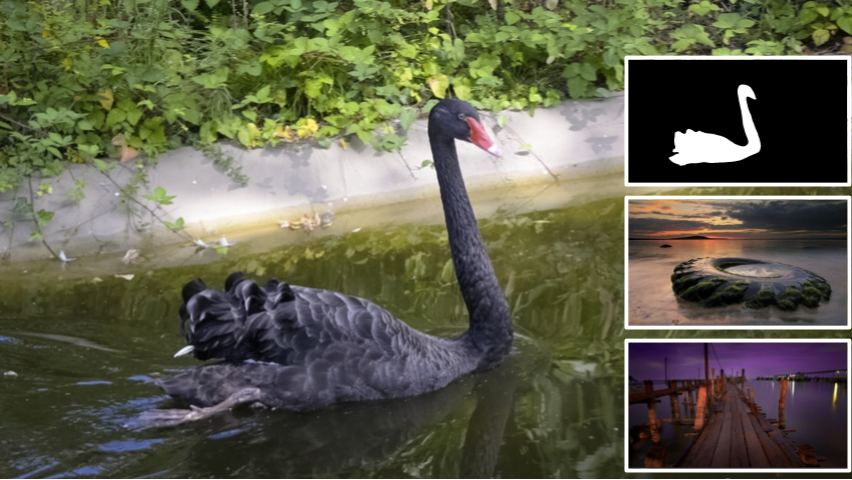} &
  \includegraphics[width = .33\linewidth,height=0.17\linewidth]{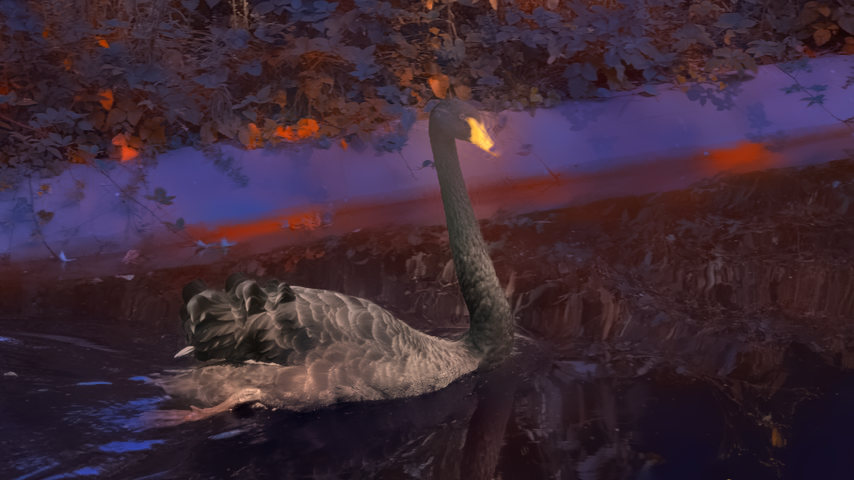} &
  \includegraphics[width = .33\linewidth,height=0.17\linewidth]{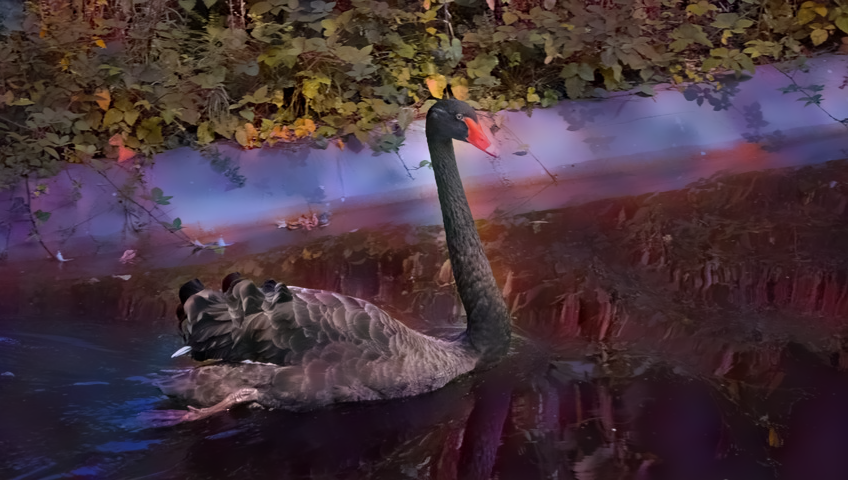} \\
  Inputs &PhotoWCT~\cite{li2018closed} &  LST~\cite{li2019learning} \\
  
  \includegraphics[width = .33\linewidth,height=0.17\linewidth]{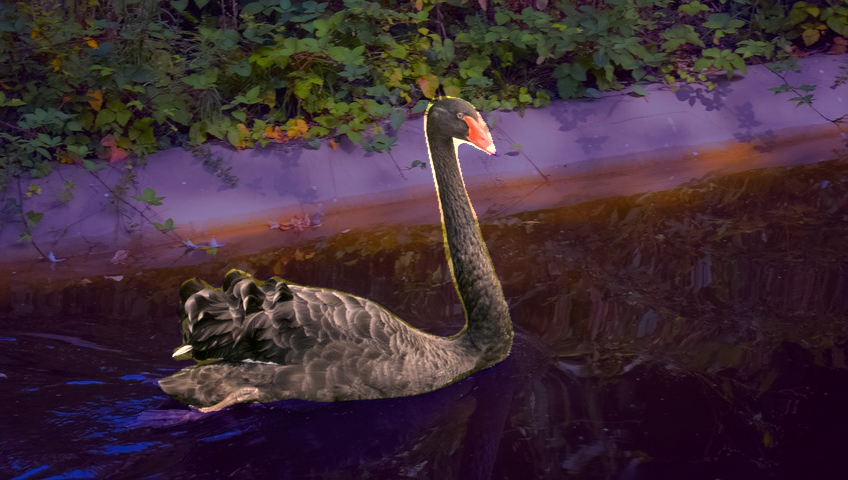} &
  \includegraphics[width = .33\linewidth,height=0.17\linewidth]{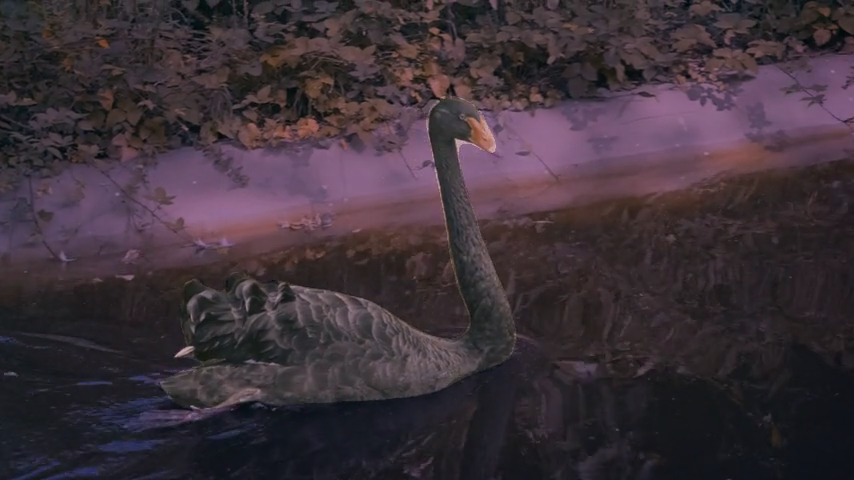} &
   \includegraphics[width = .33\linewidth,height=0.17\linewidth]{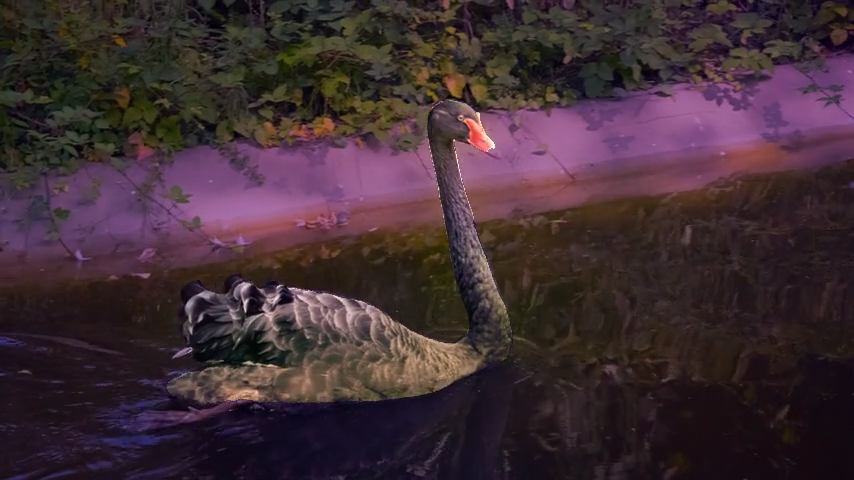} \\
   WCT\textsuperscript{2}~\cite{yoo2019photorealistic} & Xia et al.~\cite{xia2020joint} & Ours  \\

  \includegraphics[width = .32\linewidth,height=0.16\linewidth]{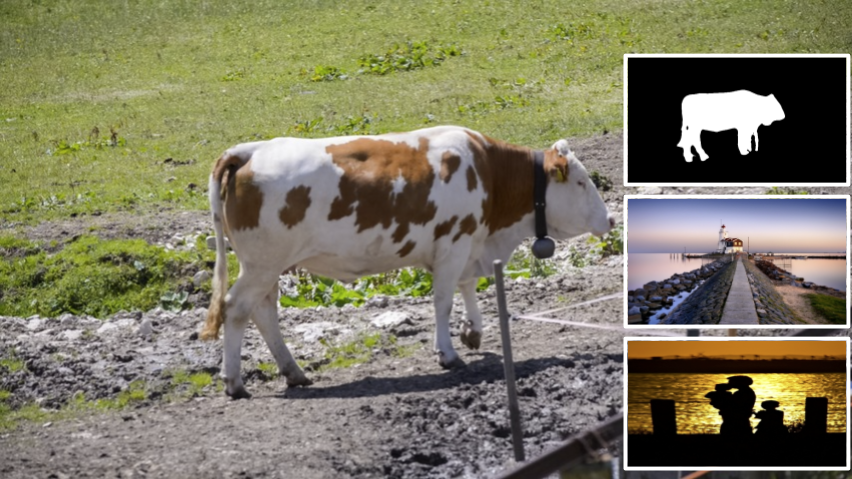} &
  \includegraphics[width = .32\linewidth,height=0.16\linewidth]{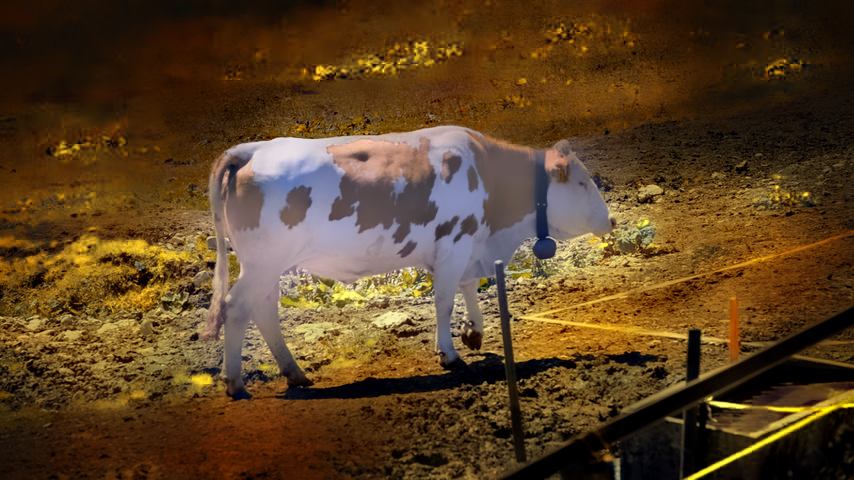} &
  \includegraphics[width = .32\linewidth,height=0.16\linewidth]{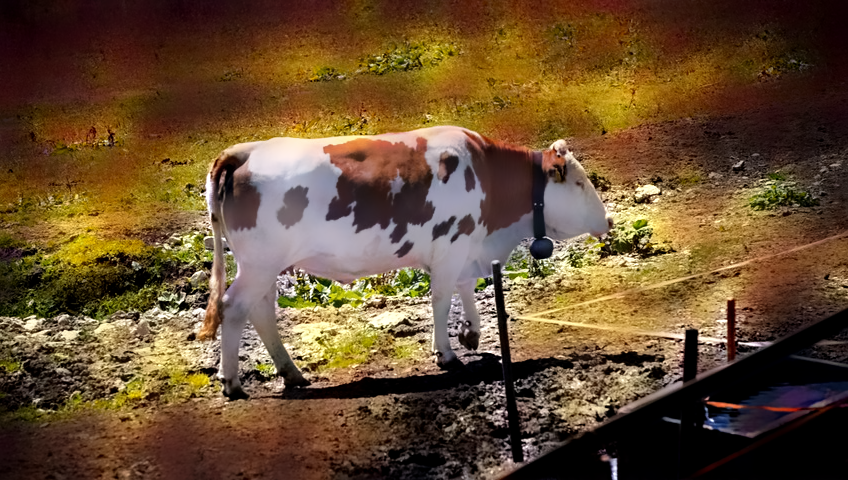} \\
  Inputs &PhotoWCT~\cite{li2018closed} &  LST~\cite{li2019learning} \\
  
  \includegraphics[width = .32\linewidth,height=0.16\linewidth]{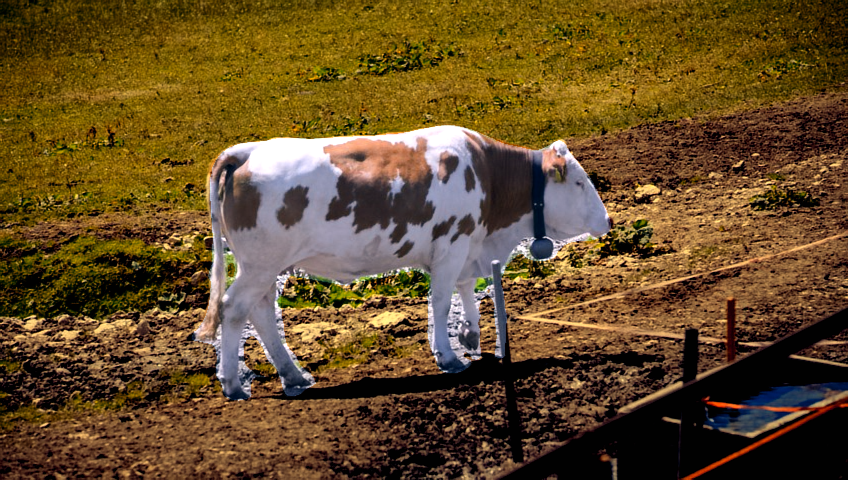} &
  \includegraphics[width = .32\linewidth,height=0.16\linewidth]{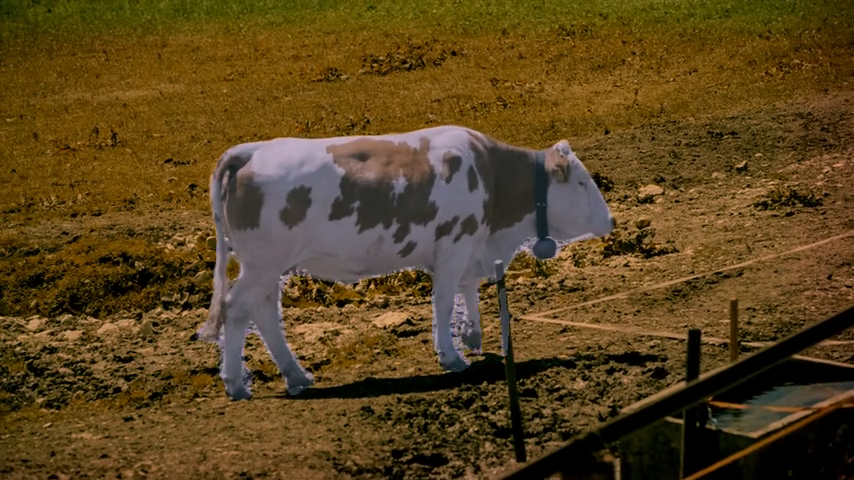} &
  \includegraphics[width = .32\linewidth,height=0.16\linewidth]{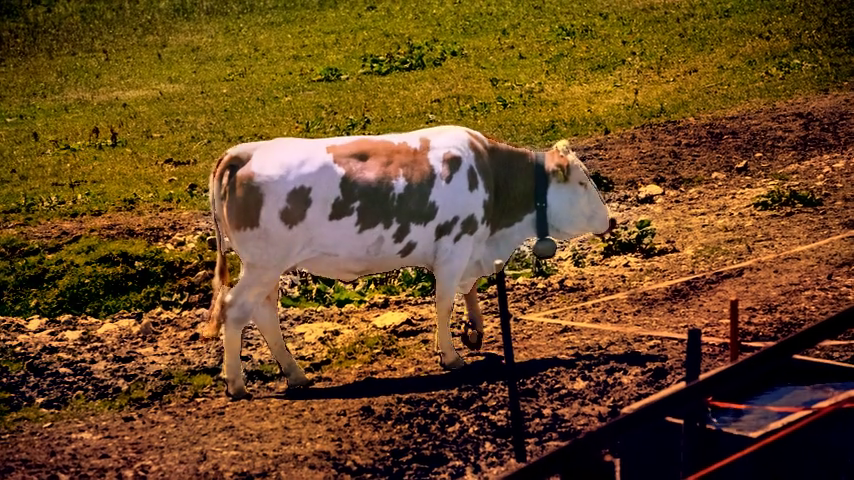} \\
   WCT\textsuperscript{2}~\cite{yoo2019photorealistic} & Xia et al.~\cite{xia2020joint} & Ours  \\
\end{tabular}
\vspace{-1mm}
\caption{\textbf{Visual comparisons.} The proposed method generates more visually pleasing stylized results than state-of-the-art photorealistic style transfer algorithms.}
\label{fig:qualitative}
\vspace{-1mm}
\end{figure*}


\subsection{Setup}
We use the DAVIS 2017 validation set~\cite{davis2017} for evaluation, which contains 30 videos with ground-truth segmentation masks and the video resolution is $480\times584$.
%
%
In our experiments, we adopt an instance segmentation method, COSNet~\cite{lu2019see}, to select a single instance as foreground.
Our system can also take multiple instances as foreground with additional user input (e.g., “control click”).

We compare our method to state-of-the-art photorealistic style transfer algorithms: PhotoWCT~\cite{li2018closed}\footnote{We use NVIDIA's latest FastPhotoStyle library, which is much faster than the speed claimed in their paper.}, LST~\cite{li2019learning}, WCT\textsuperscript{2}~\cite{yoo2019photorealistic}, and Xia et al.~\cite{xia2020joint}. 
Note that we apply the full pipeline of PhotoWCT and LST, including their own smoothing steps.
Since these approaches are intended for still image style transfer, to make the comparison more fair, we apply the temporal consistency algorithm of Lai~et~al.~\cite{lai2018learning} to as a post-process to improve their results.
As the method of Xia~et~al.~\cite{xia2020joint} 
cannot stylize local regions, we extend it by first generating two stylized outputs for foreground and background, then blending them in pixel-space using the input mask.



\subsection{Qualitative Comparisons}
\label{sec:qualitative}
Figure~\ref{fig:qualitative} provides visual comparisons on two examples.
As both PhotoWCT and LST transfer style features on the intermediate layers of a deep neural network with an encoder-decoder architecture, foreground and background styles are mixed together. Hence, the final output synthesized by the decoder contains halo artifacts at object boundaries (e.g., the swan's neck for PhotoWCT) or spatially inconsistent color (e.g., the uneven grass).
Those visual artifacts cannot be easily removed by their post-processing steps.
Therefore, the results of PhotoWCT and LST still look non-photorealistic with noticeable artifacts.
%
While WCT\textsuperscript{2} and Xia et al.~\cite{xia2020joint} do not require any post-processing, their stylized results contain visual artifacts around the object boundaries due to inaccuracies in the input object masks.
In contrast, our method produces visually pleasing results given imperfect segmentation masks.
By using our mask-enhancement and grid-space blending, we can spatially blend between substantially different styles without explicitly separating foreground and background with pixel precision.
We provide video comparisons in the supplementary material, which shows that our model generates results that are more temporally stable.



\subsection{Quantitative Results}
\label{sec:quantitative}


\Paragraph{Performance.}
We measure execution time on a single NVIDIA Tesla V100 GPU with 16 GB of RAM at various resolutions. 
For existing methods~\cite{li2019learning,li2018closed,yoo2019photorealistic,xia2020joint}, we include time spent on post-processing using Lai's method~\cite{lai2018learning} (also listed independently).
As shown in Table~\ref{tab:time}, our method is much faster than all but the method of Xia~et~al.~\cite{xia2020joint} upon which it is based.
This is due to our model's additional steps of computing a soft grid-space  mask.
Still, our model can run at 21.5~Hz on 12 megapixel inputs, while other approaches run out of memory.
%
We note that Xia~et~al.~\cite{xia2020joint} alone can run on $3000 \times 4000$ input, but the post-processing by Lai~et~al.~\cite{lai2018learning} runs out of memory at this resolution.

\begin{table}
    \centering
    \resizebox{\linewidth}{!}{
    \begin{tabular}{l|c|c|c|c}
        \toprule
        Image Size & 512 $\times$ 512 & 1024 $\times$ 1024 & 2000 $\times$ 2000 & 3000 $\times$ 4000 \\
        \midrule
        Lai et al.~\cite{lai2018learning} & 0.002$\pm$0.001s  & 0.003$\pm$0.001s &  0.008$\pm$0.003 s & OOM \\
        \midrule
        LST*~\cite{li2019learning} &   0.260$\pm$0.125s &  0.793$\pm$0.334s & OOM & OOM \\
        PhotoWCT*~\cite{li2018closed} & 0.624$\pm$0.298s &  1.548$\pm$0.359s & OOM & OOM\\
        WCT\textsuperscript{2}*~\cite{yoo2019photorealistic} &  3.904$\pm$0.280s & 6.203$\pm$0.453s & OOM & OOM \\
        Xia et al.*~\cite{xia2020joint} & 0.005$\pm$0.001s & 0.007$\pm$0.002s & 0.013$\pm$0.004s & OOM\\
        Ours & 0.039$\pm$0.001s & 0.040$\pm$0.012s & 0.043$\pm$0.013s & 0.046$\pm$0.014s \\
        \bottomrule
    \end{tabular}
    }
    \vspace{-1.em}
    \caption{\textbf{Execution time.} An asterisk ($\ast$) denotes that the technique of Lai~et~al.~\cite{lai2018learning} is applied. OOM indicates out of memory at the inference time. }
    \label{tab:time}
\vspace{-1.em}
\end{table}


\Paragraph{User study.}
As style transfer fidelity is inherently a subjective matter of taste, we conduct a user study to evaluate user's preferences.
We adopt paired comparisons~\cite{rubinstein2010comparative,lai2016comparative}, where users are asked to choose the better result in a pair shown side-by-side.
In each test, we also provide the input content video, object segmentation mask, foreground, and background style images as reference.
The participant is asked to answer the following questions:
\begin{compactenum}
\item Which video has a more faithful stylization?
\item Which video has fewer visual artifacts?
\item Which video is more temporally stable?
\end{compactenum}
In total, we recruit 30 participants, where each participant evaluates 20 sets of videos.
While the results are shuffled randomly, we ensure that all the methods are compared the same number of times.

Table~\ref{tab:user_study} shows the percentage that our result is preferred over the other.
Overall, our method is selected on more than $60\%$ of comparisons for all three questions, demonstrating that our results have a more faithful stylization, fewer visual artifacts, and better temporal stability.
\ignorethis{We note that WCT\textsuperscript{2}~\cite{yoo2019photorealistic} achieves comparable results in terms of stylization ($58.8\%$), which shares a similar observation to the evaluation in Xia et al.~\cite{xia2020joint}.}

\begin{table}
    \centering
    \resizebox{\linewidth}{!}{
    \begin{tabular}{l|c|c|c}
        \toprule
        & Better & Fewer visual & Better temporal \\
        & stylization & artifacts & stability \\
        \midrule
        Ours vs. LST*~\cite{li2019learning} &
        60.6$\pm$7.5$\%$ & 68.5$\pm$7.2$\%$ & 66.1$\pm$7.3$\%$ \\
        Ours vs. PhotoWCT*~\cite{li2018closed} &
        53.9$\pm$7.7$\%$ & 80.6$\pm$6.1$\%$ & 73.9$\pm$6.8$\%$ \\
        Ours vs. WCT\textsuperscript{2}*~\cite{yoo2019photorealistic} &
        57.6$\pm$7.6$\%$ & 79.4$\pm$6.2$\%$ & 73.9$\pm$6.8$\%$ \\
        Ours vs. Xia et al.*~\cite{xia2020joint} &
        69.1$\pm$7.1$\%$ & 60.6$\pm$7.5$\%$ & 57.6$\pm$7.6$\%$ \\
        \midrule
        Average &
        60.3$\pm$3.7$\%$ & 72.3$\pm$3.4$\%$ & 67.9$\pm$3.6$\%$ \\
        \bottomrule
    \end{tabular}
    }
    \vspace{-1mm}
    \caption{\textbf{Results of user study.} An asterisk ($\ast$) denotes that the technique of Lai~et~al.~\cite{lai2018learning} is applied. Overall, our results are preferred by more than $60\%$ of users.}
    \label{tab:user_study}
\vspace{-1.em}
\end{table}

\begin{figure}
 \begin{center}
\includegraphics[width=\linewidth]{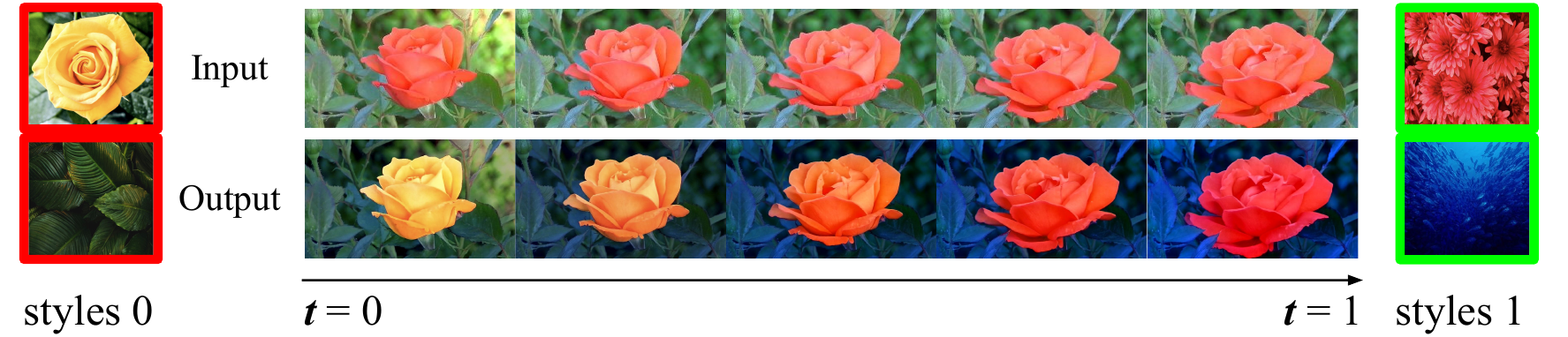}
\end{center}
\vspace{-1.em}
\caption{\textbf{Style transition.} Our model can smoothly transit between styles in time by interpolating the bilateral grids.}
\label{fig:Transition}
\vspace{-1.em}
\end{figure}

\begin{figure*}
\centering
\footnotesize
\begin{tabular}{c@{\hspace{0.005\linewidth}}c@{\hspace{0.005\linewidth}}c@{\hspace{0.005\linewidth}}c@{\hspace{0.005\linewidth}}c}

  \includegraphics[width = .19\linewidth,height=.22\linewidth]{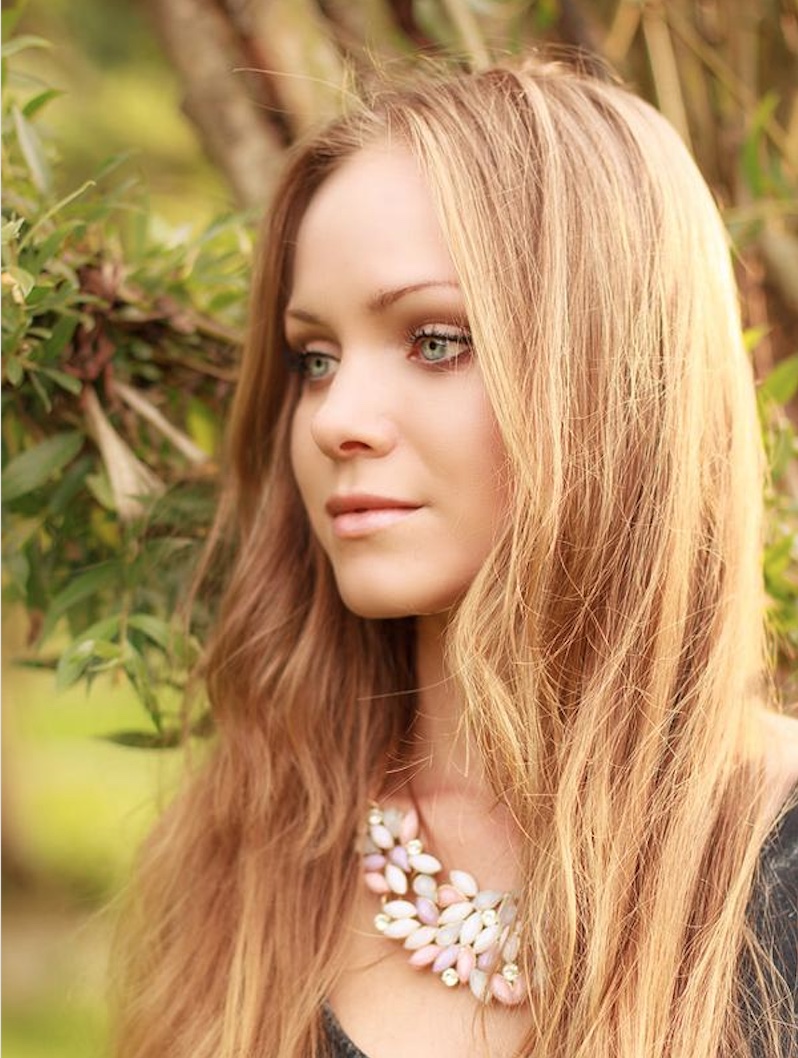}&
  \includegraphics[width = .19\linewidth,height=.22\linewidth]{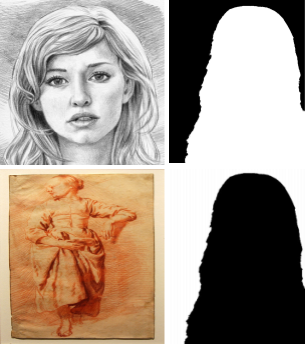}&
  \includegraphics[width = .19\linewidth,height=.22\linewidth]{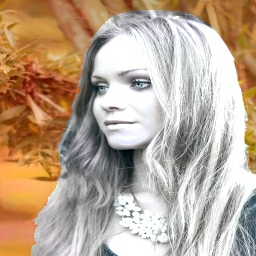}&
  \includegraphics[width = .19\linewidth,height=.22\linewidth]{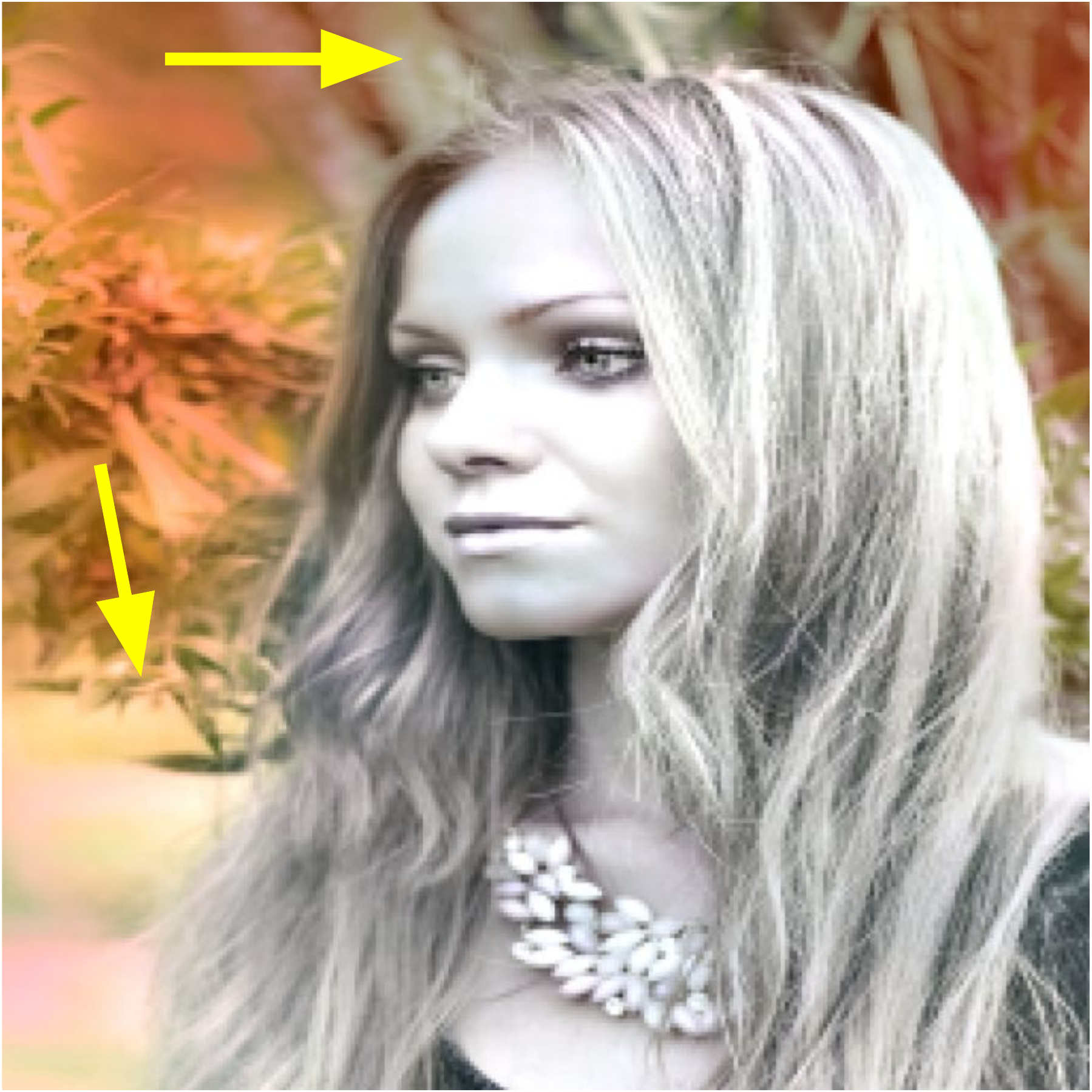}
  &
  \includegraphics[width = .19\linewidth,height=.22\linewidth]{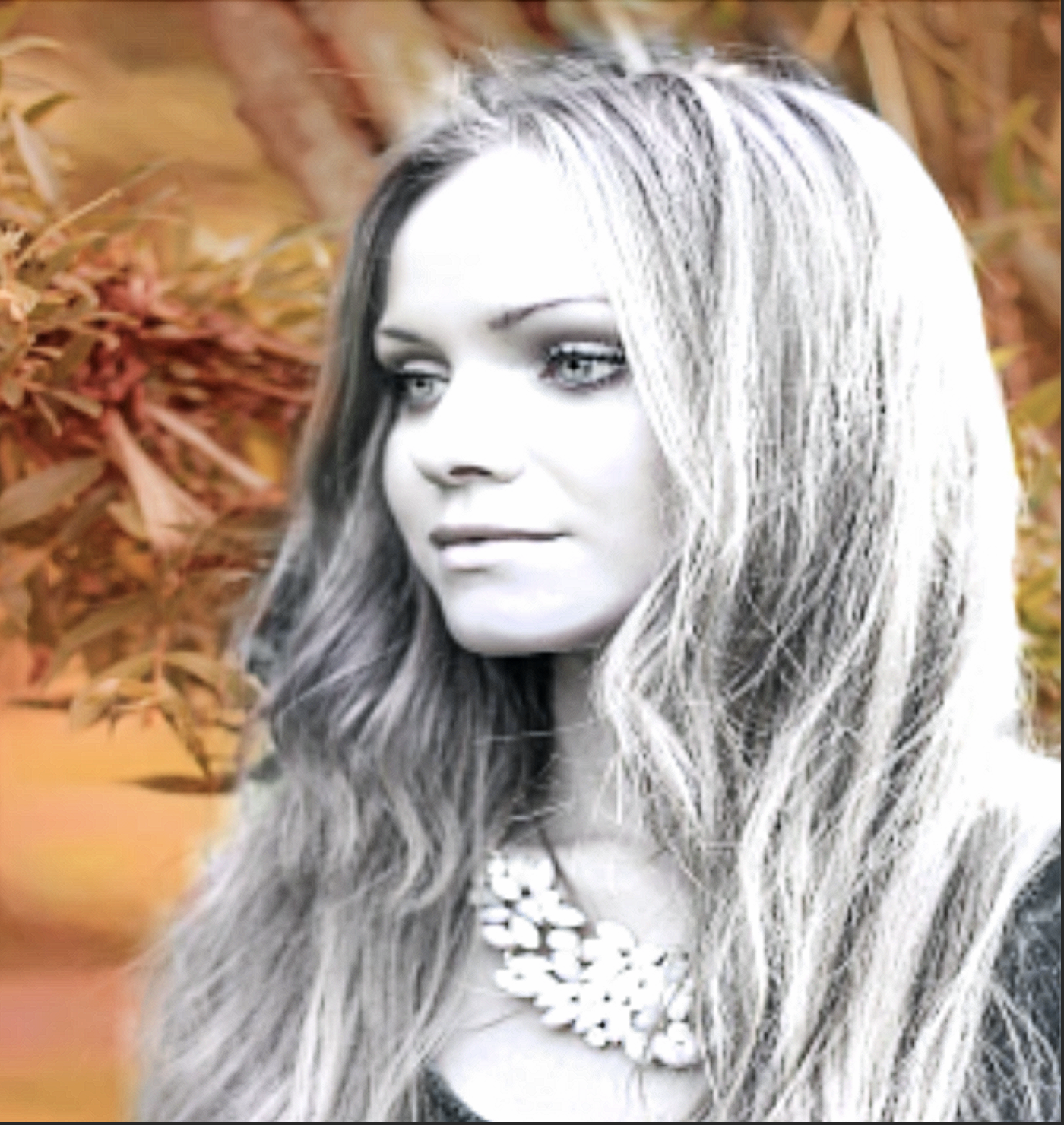}\\
 
 (a) Content & (b) Styles and masks & (c) Pixel-space blending & (d) Grid-space blending  & (e) Grid-space blending \\
 & & &  w/o SA-AdaIN &  w/ SA-AdaIN
\end{tabular}
\vspace{-1.em}
\caption{\textbf{Effect of grid-space blending and SA-AdaIN.}
Blending in the pixel-space results in unnatural object boundaries due to the hard edge in the mask.
Our grid-space blending can render smoother boundaries, but the foreground style bleeds to the background (the yellow arrows in (d)).
With the proposed SA-AdaIN, our model can separate the foreground and background styles well and render high-quality object boundaries.}
\label{fig:pixel_stitching}
\vspace{-1.em}
\end{figure*}

\subsection{Style Transition}

The style and color tone in a film can change over time, connoting a gradual shift in mood.
Our method can easily transition between different styles in time by interpolating between predicted grids using an appropriate time-dependent weighting function (e.g., a user-selected spline).
Figure~\ref{fig:Transition} samples a few frames from a result where we transfer different styles onto foreground and background while smoothly transitioning in time.
The full video result is shown in the supplementary file.


\subsection{Ablation Studies}
\label{sec:ablation}

\Paragraph{Grid-space blending and SA-AdaIN.}
%
%
As input object masks are often noisy, blending in pixel space often results in visible artifacts around object boundaries (Figure~\ref{fig:pixel_stitching}(c)).
To render natural object boundaries, one may need to apply image matting~\cite{xu2017deep} to feather the mask.
But matting is both computationally expensive and may introduce its own artifacts.
In contrast, our method learns how to blend the foreground and background style transfer transforms, sidestepping the need for an accurate full-resolution mask (Figure~\ref{fig:pixel_stitching}(e)).
%
%
Furthermore, by adopting the SA-AdaIN (Equation~\eqref{eq:sa-adain}, we improve the separation of foreground and background styles better and avoid the classic halo artifacts as shown in Figure~\ref{fig:pixel_stitching}(d).

\begin{figure}
\centering
\footnotesize
\begin{tabular}{c@{\hspace{0.005\linewidth}}c@{\hspace{0.005\linewidth}}c}

  \includegraphics[width = .33\linewidth,height=.18\linewidth]{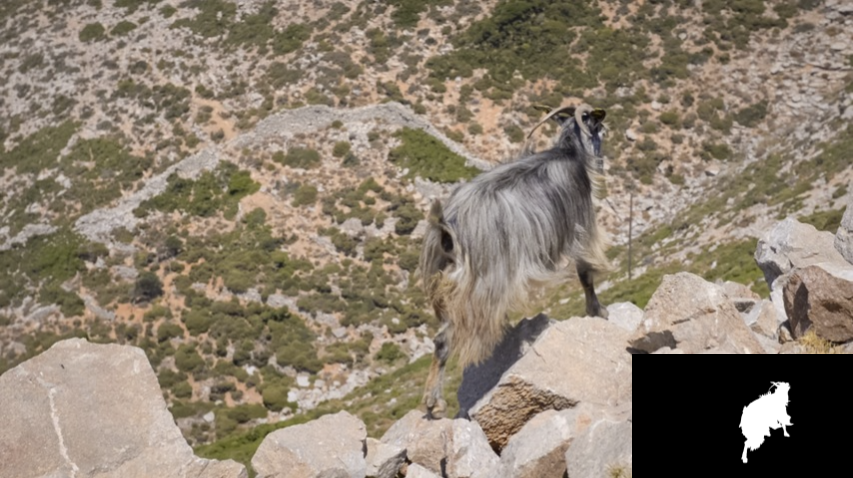} &
  \includegraphics[width = .33\linewidth,height=.18\linewidth]{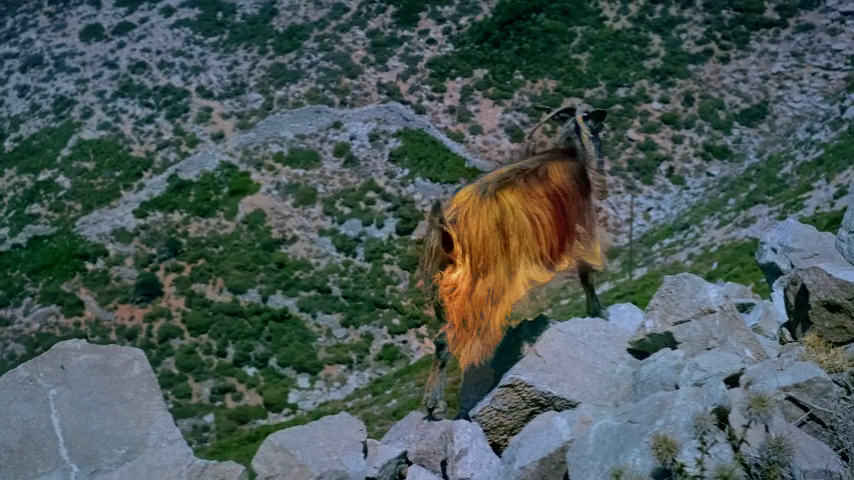} &
  \includegraphics[width = .33\linewidth,height=.18\linewidth]{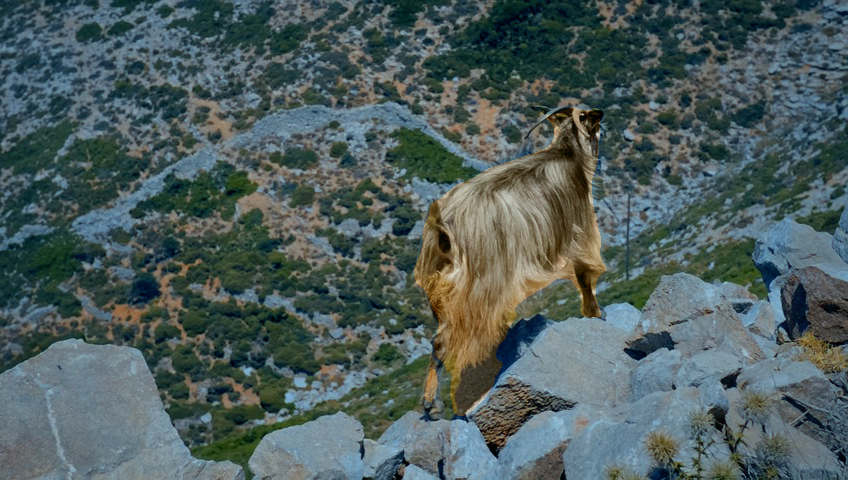}  \\
 
   \includegraphics[width = .33\linewidth,height=.18\linewidth]{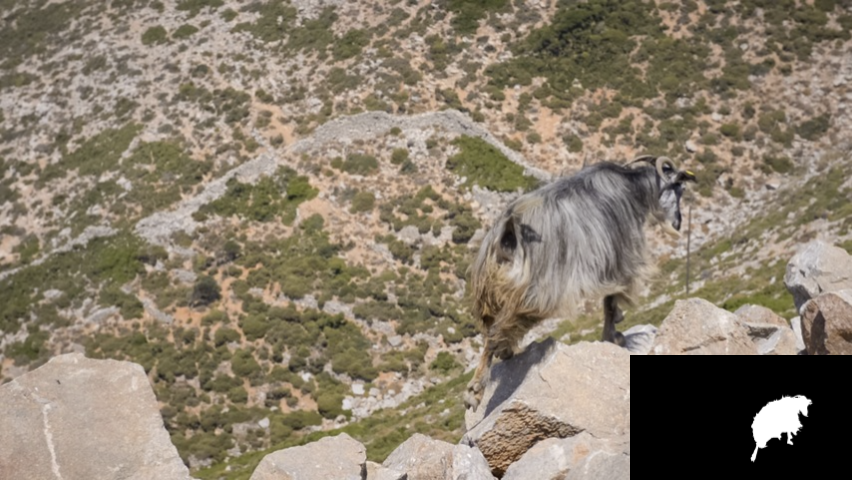} &
  \includegraphics[width = .33\linewidth,height=.18\linewidth]{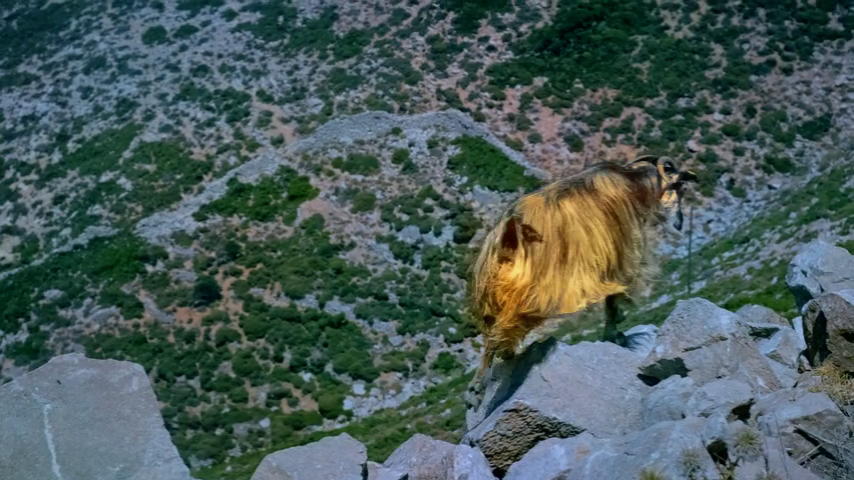} &
  \includegraphics[width = .33\linewidth,height=.18\linewidth]{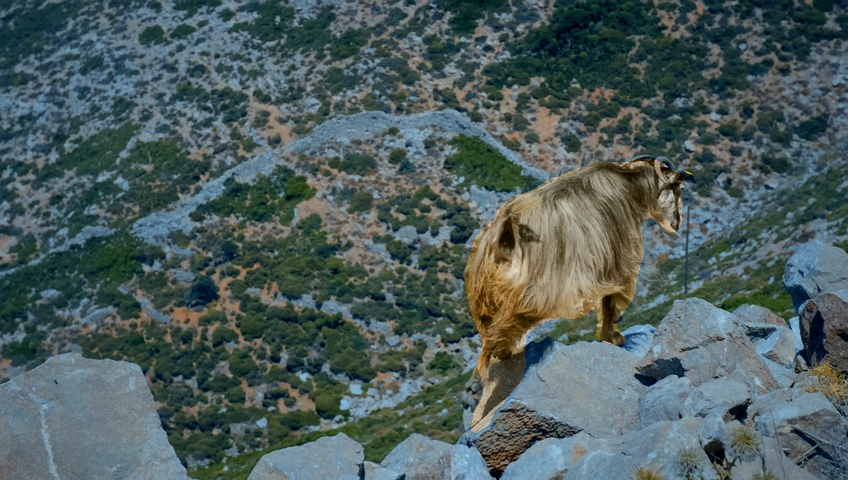}  \\
  Inputs  &  w/o TC-AdaIN & Final results
\end{tabular}
\vspace{-1mm}
\caption{\textbf{Effect of TC-AdaIN.} Without propagating intermediate features with TC-AdaIN, the stylized outputs have a significant color shifting within 8 frames in this example.}
\label{fig:tc-adain_ablation}
\vspace{-1mm}
\end{figure}

\Paragraph{Temporal consistency.}
To quantitatively evaluate the temporal consistency of the stylized videos, we measure the warping error~\eqref{eq:st_loss} and temporal change consistency (TCC)~\cite{zhang2019exploiting}.
We compare the per-frame results of PhotoWCT~\cite{li2018closed}, LST~\cite{li2019learning}, WCT\textsuperscript{2}~\cite{yoo2019photorealistic}, and Xia~et~al.~\cite{xia2020joint}, both with and without applying the temporal smoothing technique of Lai~et~al.~\cite{lai2018learning}.
We also train two modified versions of our model: one without the temporal loss and one without TC-AdaIN.
As shown in Table~\ref{tab:temporal_consistency}, our full model achieves the lowest warping error and the highest TCC, demonstrating that our method achieves the most stable results.
Figure~\ref{fig:tc-adain_ablation} also shows a visual comparison, where the results without the TC-AdaIN have a visible color shift.

%
%

%


\begin{table}
    \footnotesize
    \centering
    \resizebox{0.9\linewidth}{!}{
    \begin{tabular}{l|c|c|c|c}
        \toprule
        Model & 
        Warping error ($\downarrow$) & TCC ($\uparrow$)\\
        \midrule
        PhotoWCT~\cite{li2018closed} & 0.00124 & 0.632\\
        LST~\cite{li2019learning} & 0.00129 & 0.534\\
        WCT\textsuperscript{2}~\cite{yoo2019photorealistic} & 0.00118 & 0.566\\
        Xia et al.~\cite{xia2020joint} & 0.00098 & 0.607\\
        \midrule
        PhotoWCT~\cite{li2018closed} + Lai et al.~\cite{lai2018learning} & 0.00099 & 0.643\\
        LST~\cite{li2019learning} + Lai et al.~\cite{lai2018learning} & 0.00104 & 0.538\\
        WCT\textsuperscript{2}~\cite{yoo2019photorealistic} + Lai et al.~\cite{lai2018learning} & 0.00110 & 0.576\\
        Xia et al.~\cite{xia2020joint} + Lai et al.~\cite{lai2018learning} & 0.00093 & 0.660\\
        \midrule
        Ours w/o $\mathcal{L}_{t}$ & 0.00091 & 0.673\\ 
        Ours w/o TC-AdaIN & 0.00094 & 0.659\\ 
        \textbf{Ours} & \textbf{0.00090} & \textbf{0.688}\\ 
        \bottomrule
    \end{tabular}
    }
    \vspace{-1mm}
    \caption{\textbf{Temporal consistency evaluation.} Our full model achieves the lowest warping error and the highest TCC against the existing algorithms and our own variations.}
    \label{tab:temporal_consistency}
    \vspace{-1mm}
\end{table}

\section{Conclusion}

In this work, we propose a novel algorithm for photorealistic video style transfer.
Our algorithm is able to 1) generate spatially and temporally coherent stylized results throught the proposed spatiotemporal feature transfer layer, 
2) account for imperfect input masks by learning a mask enhancement network and blending the color transform coefficients in a low-resolution grid space,
and 3) achieve real-time performance by adopting an efficient bilateral learning framework.
%
%
We demonstrate the performance of the proposed method on a wide variety of videos and styles.
Finally, a human subject study shows that our method achieves faithful stylization, higher visual quality, and better temporal coherence against existing photorealistic style transfer approaches.

\section{Algorithm Details}

\subsection{Network architectures}
\begin{figure*}
\centering
\includegraphics[width=1.0\linewidth]{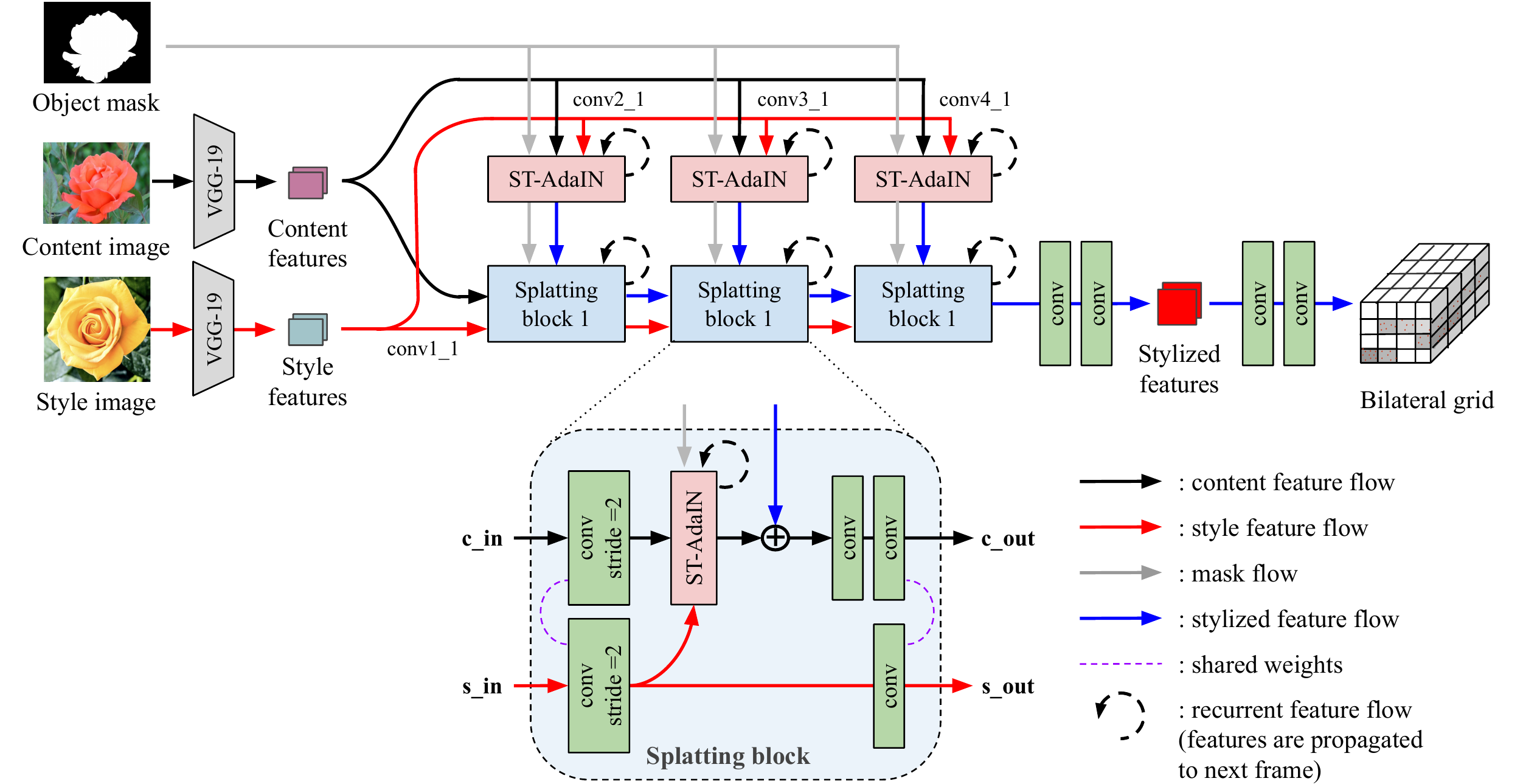}
\caption{Detailed architecture of our spatiotemporal feature transfer and grid prediction layers.}
\label{fig:st_adain}
\end{figure*}

We provide the architectural details of our mask enhancement network, guidance learner, spatiotemporal feature transfer, and grid learner in Table~\ref{tbl:layers_specifics}.
As shown in Figure~\ref{fig:st_adain}, we extract the VGG-19 features from four scales (\textsc{conv1\_1}, \textsc{conv2\_1}, \textsc{conv3\_1}, and \textsc{conv4\_1}) and match the multi-resolution feature statistics via three splatting blocks~\cite{xia2020joint}.
Note we remove the global scene summary path in Xia et al. ~\cite{xia2020joint} as we aim to transfer style between local regions, where the global features will not help.
%
%
%
In each splatting block, the weights of the first and the last convolutional layers are shared for the content and style feature paths.
The output of each ST-AdaIN layer is propagated to the next frame for blending the temporal information.
After the feature transferring, we apply two convolutional layers to predict the bilateral grids.

\begin{table*}
\centering
\resizebox{\linewidth}{!}{
\begin{tabular}{c | ccc| cc|ccccccccccc | cc }
\toprule
 & \multicolumn{3}{c|}{Mask } & \multicolumn{2}{|c|}{Guidance } & \multicolumn{11}{|c|}{Spatiotemporal } & \multicolumn{2}{|c}{Grid } \\
 & \multicolumn{3}{c|}{ Enhancement} & \multicolumn{2}{|c|}{Learner } & \multicolumn{11}{|c|}{Feature Transfer} & \multicolumn{2}{|c}{Prediction} \\ \midrule
               & $M^1$ & $M^2$ & $M^3$&   $G^1$  & $G^2$    & $S_1^1$ & $S_1^2$ & $S_1^3$ & $S_2^1$ & $S_2^2$ & $S_2^3$ & $S_3^1$ & $S_3^2$ & $S_3^3$ & $L^1$ & $L^2$      & $F$  & $\Gamma$ \\ \midrule
type           & conv   & conv   & conv      & conv  & conv      & conv     & conv     & conv     & conv     & conv     & conv     & conv     & conv     & conv     & conv   & conv        & conv  & conv  \\
stride         & 1     & 1     & 1        & 1    & 1        & 2       & 1       & 1       & 2       & 1       & 1       & 2       & 1       & 1       & 1     & 1          & 1    & 1  \\
kernel size    & 3     & 3     & 3        & 3    & 3        & 3       & 3       & 3       & 3       & 3       & 3       & 3       & 3       & 3       & 3     & 3          & 3    & 1  \\
spatial size   & 256   & 256   & 256      & 256  & 256      & 128     & 128     & 128     & 64      & 64      & 64      & 32      & 32      & 32      & 16    & 16         & 16   & 16 \\
channels       & 16    & 8     & 1        & 16   & 1        & 8       & 8       & 8       & 16      & 16      & 16      & 32      & 32      & 32      & 64    & 64         & 64   & 96 \\ 
activation     & -     & -     & sigmoid  & -    & sigmoid  & relu    & relu    & relu    & relu    & relu    & relu    & relu    & relu    & relu    & relu    & relu     & relu & -  \\ \bottomrule
\end{tabular}
}
\caption{Detailed configuration of each module. 
}
\label{tbl:layers_specifics}
\end{table*}

\subsection{Soft grid mask}

In Algorithm~\ref{alg:grid_mask}, we provide the detailed steps to compute the soft grid mask for blending the foreground and background grids.

\begin{algorithm}
\SetAlgoLined
\KwInput{Learned guide map $\mathbf{z}$, pixel mask $\mathbf{{M}_{pxl}}$, image size $(w,h)$, grid size $(W,H,D)$}
\KwOutput{Soft grid mask $\mathbf{{M}_{grid}}$}
 Initialize grid mask: ${M}_{grid} = zeros(W,H,D)$\;
 $\mathbf{z}_D = floor((\mathbf{z}\cdot \mathbf{{M}_{pxl}}) \times D)$ \;
 $s_w, s_h = w/W, h/H$ \;
 
 \For{$x \gets 1$ to $W$ and $y \gets 1$ to $H$}{
    $patch = \mathbf{z}_D[x \times s_w:(x+1) \times s_w, h \times s_h:(y+1) \times y_h]$ \;
    ${M}_{grid}[x,y,:] \gets sum(patch>0)$\;
    \For{$d \gets 1$ to $D$}{
        \If{$d$ in $patch$}{
           ${M}_{grid}[w,h,d] \gets sum(patch==d)$\;
        }
    }
}
Normalize grid mask: ${M}_{grid} \gets {M}_{grid}/(s_w \times s_h)$\;
\caption{Compute Soft Grid Mask}
\label{alg:grid_mask}
\end{algorithm}

\subsection{Implementation details}
\label{subsec:training}

We implement our model in Tensorflow~\cite{abadi2016tensorflow}.
For training, we use the training set from DAVIS 2017~\cite{davis2017} which contains 60 videos and each video has 70 frames in average. 
We optimize the model using Adam~\cite{kingma2015adam} with hyperparameters $\alpha=10^{-4}, \beta_1=0.9, \beta_2=0.999, \epsilon=10^{-8}$, and a batch size of $2$ video clips.
At each iteration, we take five consecutive frames as a clip and randomly crop the original frames into a resolution of $256 \times 256$. 
To train the mask enhancement network, we generate noisy input masks by applying erosion and dilation with random kernel sizes to the ground-truth object masks.
Since mask enhancement is a largely orthogonal task, we first pre-train this module for 20000 iterations and freeze the weights.
%
We then train the remainder of the network end-to-end for $90000$ iterations.
By limiting the ``full-res'' input to also be $256 \times 256$, we can significantly reduce training time, which takes about two days on a single NVIDIA Tesla V100 GPU with 16 GB RAM.
At inference time, the trained model can be applied to arbitrary input resolution, because of the usage of bilateral grid. 


\section{Grid Sub-sampling}

\begin{figure}
\centering
\includegraphics[width=1.0\linewidth]{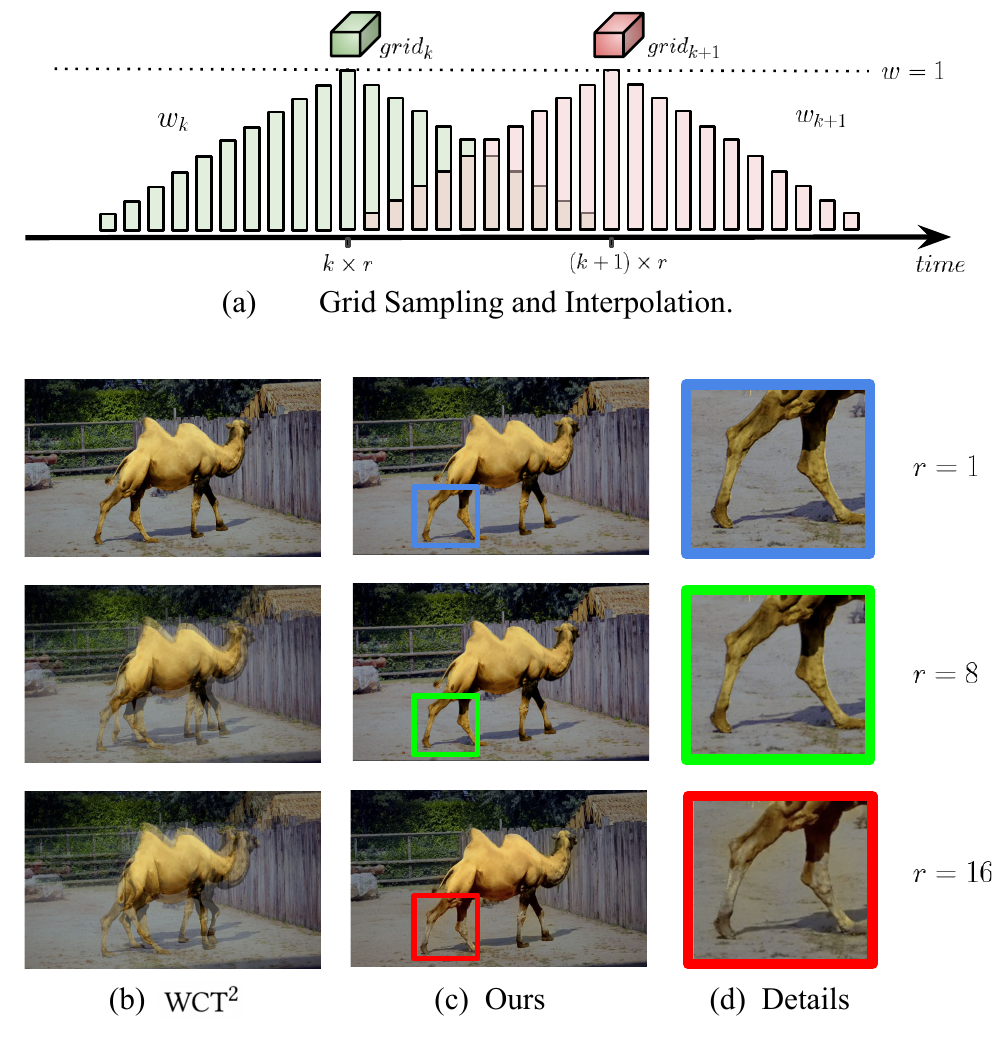}
\caption{Our method can be sped up by subsampling the bilateral grids in the temporal space while achieving similar visual quality.}
\label{fig:grid_sampling_ablat}
\end{figure}

The proposed method can be further sped up by sub-sampling the bilateral grids in the temporal domain.
As the nearby frames typically have similar color, the bilateral grids can also be shared to render the stylized result.
Specifically, we can generate the bilateral grid for every $r$ frames and estimate the intermediate grids by a linear interpolation, as shown in Figure~\ref{fig:grid_sampling_ablat}(a).
In Figure~\ref{fig:grid_sampling_ablat}(c) and (d), we show that our method can still render high-quality results up to a sub-sampling rate of $r = 8$.
For $r = 16$, the estimated grid may have a larger spatial mismatch with the content, resulting in undesired visual artifacts.
Such a temporal sub-sampling strategy is suitable for the proposed method and Xia et al.~\cite{xia2020joint}.
Other approaches, e.g., WCT\textsuperscript{2}~\cite{yoo2019photorealistic}, is not able to generate reasonable results by simply interpolating the output frames, as shown in Figure~\ref{fig:grid_sampling_ablat}(b).

By utilizing such a grid sub-sampling strategy, we can reduce the computational cost and speed up the processing time during inference.
As shown in Table~\ref{tab:time_rate8}, our model with sub-sampling rate $r = 8$ is about $8\times$ faster at four different image resolutions.
The video comparisons are provided in the supplementary videos $grid\_sampling.mp4$ in the $\textbf{video\_mp4}$ folder.

\begin{table}[ht!]
    \centering
    \resizebox{\linewidth}{!}{
    \begin{tabular}{l|c|c|c|c}
        \toprule
        Image Size & 512 $\times$ 512 & 1024 $\times$ 1024 & 2000 $\times$ 2000 & 3000 $\times$ 4000 \\
        \midrule
        Lai et al.~\cite{lai2018learning} & 0.0024s  & 0.0031s &  0.0073s & OOM \\
        \midrule
        LST*~\cite{li2019learning} &  0.2753s &  0.8365s & OOM & OOM \\
        PhotoWCT*~\cite{li2018closed} & 0.6366s &  1.5185s & OOM & OOM\\
        WCT\textsuperscript{2}*~\cite{yoo2019photorealistic} &  3.8599s & 6.1375s & OOM & OOM \\
        Xia et al.*~\cite{xia2020joint} & 0.0058s & 0.0068s & 0.0117s & OOM \\
        Ours ($r = 1$) & 0.0378s & 0.0380s & 0.0414s & 0.0464s \\
        Ours ($r = 8$) & 0.0048s & 0.0049s & 0.0052s & 0.0058s \\
        \bottomrule
    \end{tabular}
    }
    \caption{Execution time. An asterisk denotes that the technique of Lai~et ~al.~\cite{lai2018learning} was applied to improving temporal stability. OOM indicates out of memory.}
    \label{tab:time_rate8}
\end{table}



\section{More Qualitative Comparisons}

\subsection{Mask refinement}
\begin{figure*}
\centering
\footnotesize
\begin{tabular}{c@{\hspace{0.005\linewidth}}c@{\hspace{0.005\linewidth}}c@{\hspace{0.005\linewidth}}c}
  \includegraphics[width = .24\linewidth,height=.16\linewidth]{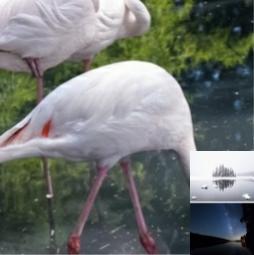} &
  \includegraphics[width = .24\linewidth,height=.16\linewidth]{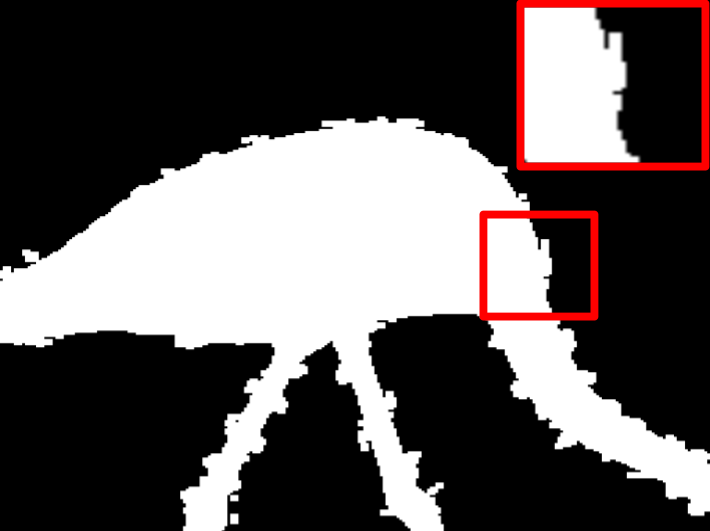} &
  \includegraphics[width = .24\linewidth,height=.16\linewidth]{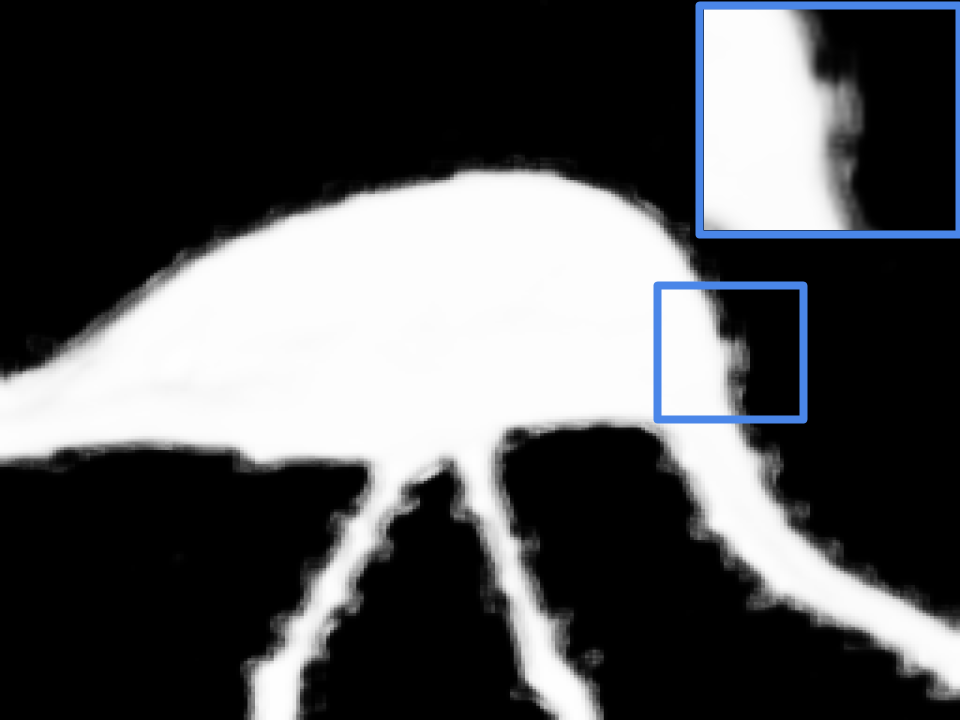} &
  \includegraphics[width = .24\linewidth,height=.16\linewidth]{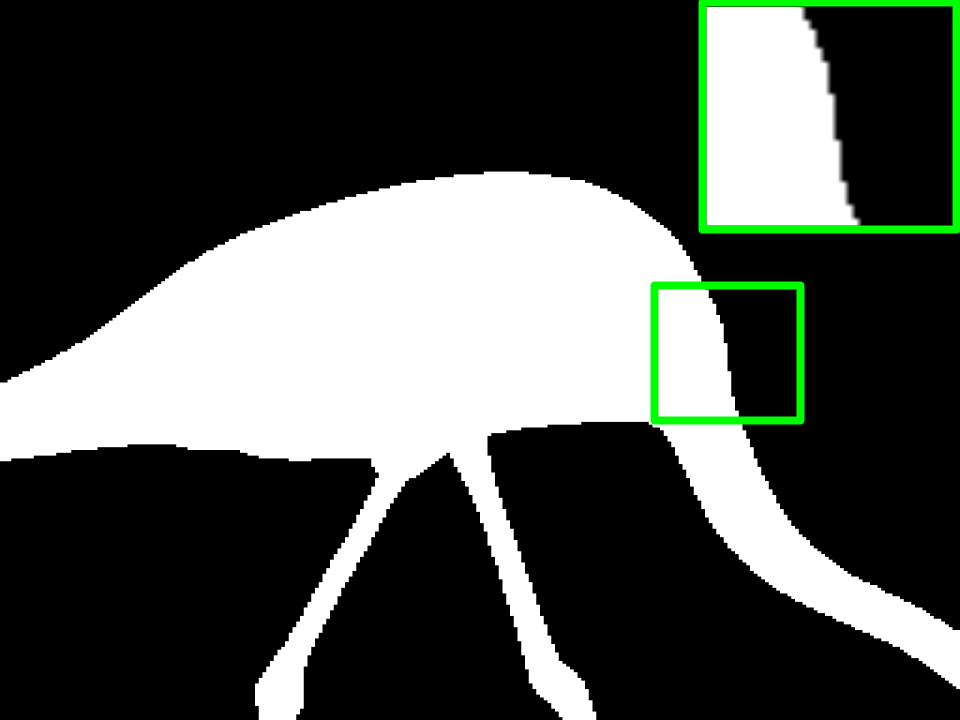} 
  \\
  (a) Inputs & (b) Input mask & (c) Enhanced mask & (d) GT mask \\
  \includegraphics[width = .24\linewidth,height=.16\linewidth]{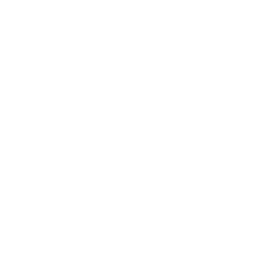} &
  \includegraphics[width = .24\linewidth,height=.16\linewidth]{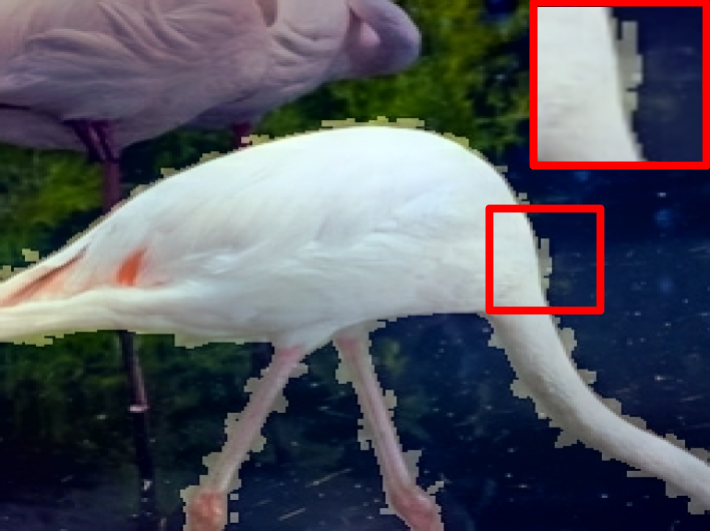} &
  \includegraphics[width = .24\linewidth,height=.16\linewidth]{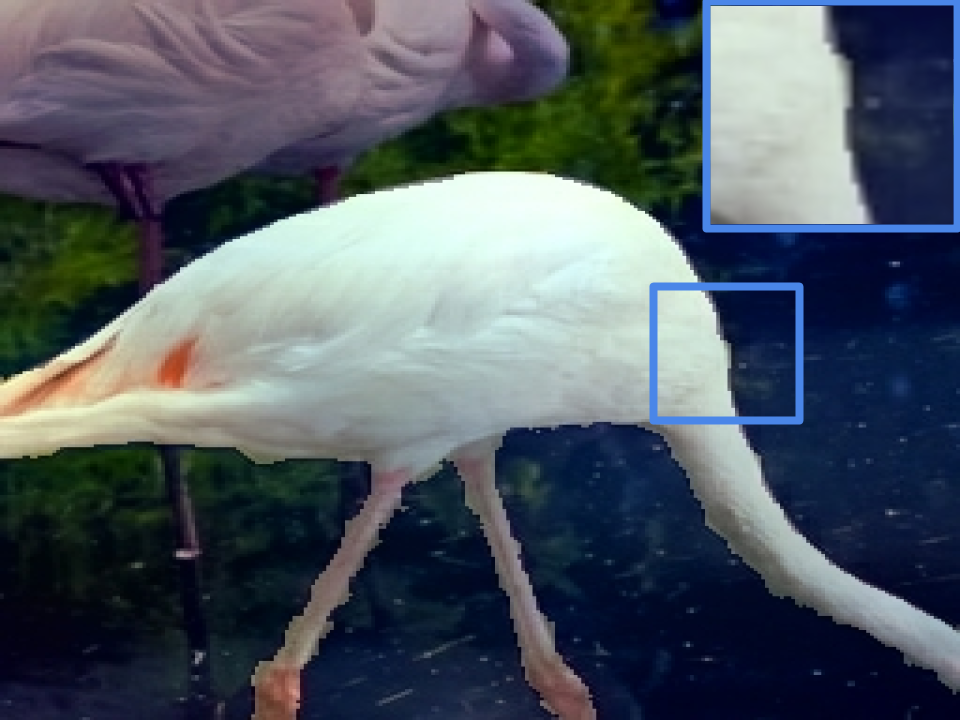} &
  \includegraphics[width = .24\linewidth,height=.16\linewidth]{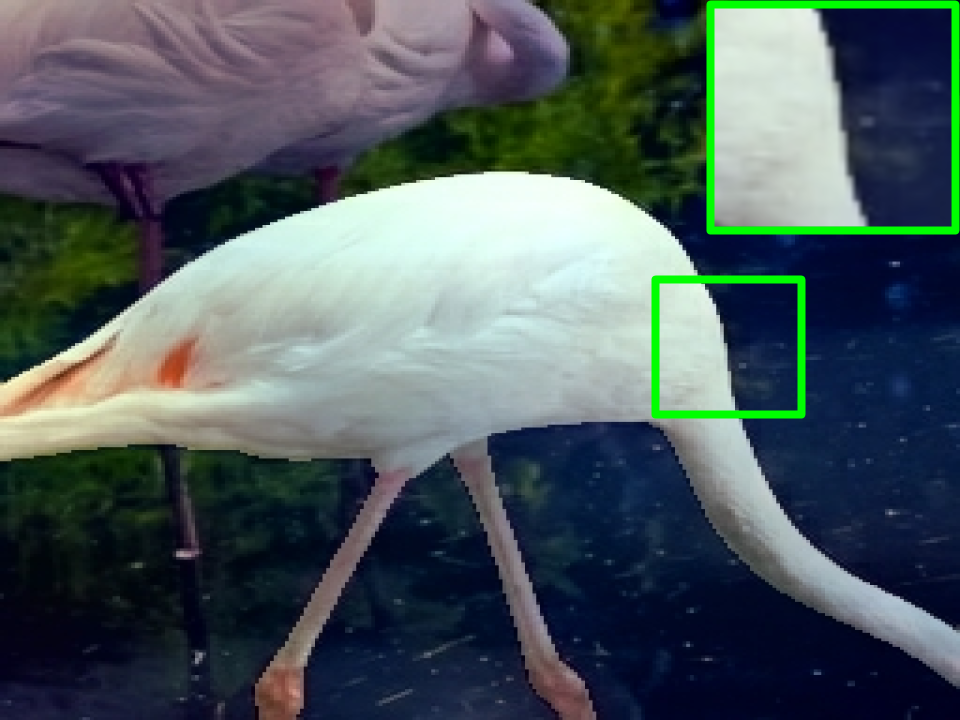}    \\
  & (e) Result with & (f) Result with & (g) Result with \\
  &  input mask & enhanced mask & GT mask \\
\end{tabular}
\vspace{-1.em}
\caption{Effect of mask enhancement. When the input object mask is too noisy, the stylized results may still contain visible artifacts on the boundaries. Mask enhancement lets us render high-quality boundaries visually comparable to the result using the ground truth mask.}
\label{fig:mask_enhance}
\vspace{-1.em}
\end{figure*}

%
%
While our grid-space blending can handle imperfect input masks, visible artifacts may remain if the masks are too noisy, as shown in Figure~\ref{fig:mask_enhance}(b) and (e).
Our mask enhancement network significantly improves mask boundaries (Figure~\ref{fig:mask_enhance}(c)), and our rendered result (Figure~\ref{fig:mask_enhance}(f)) is visually comparable to a rendering using the ground truth mask (Figure~\ref{fig:mask_enhance}(g)).
%

\subsection{Anti-distortion module}

Recent photorealistic style transfer methods~\cite{li2018closed, li2019learning, yoo2019photorealistic} are based on encoder-decoder architecture, which often cause spatial distortions or unrealistic visual artifacts when reconstructing an image from the low-resolution deep features.
Therefore, extra smoothing steps or modules are required to minimize those spatial distortions. 
PhotoWCT~\cite{li2018closed} optimizes a matting affinity to ensures spatially consistent stylization. 
Similarly, LST~\cite{li2019learning} applies a spatial propagation network (SPN)~\cite{liu2017learning} on the reconstructed images to smooth the results. 
On other other hand, WCT\textsuperscript{2}~\cite{yoo2019photorealistic} replaced max-pooling layers with wavelet pooling where the high frequency components are skipped to the decoder directly so that all edges and corners are preserved.

\begin{figure*}
\centering
\begin{tabular}{c@{\hspace{0.005\linewidth}}c@{\hspace{0.005\linewidth}}c@{\hspace{0.005\linewidth}}c}

  \includegraphics[width = .33\linewidth,height=.2\linewidth]{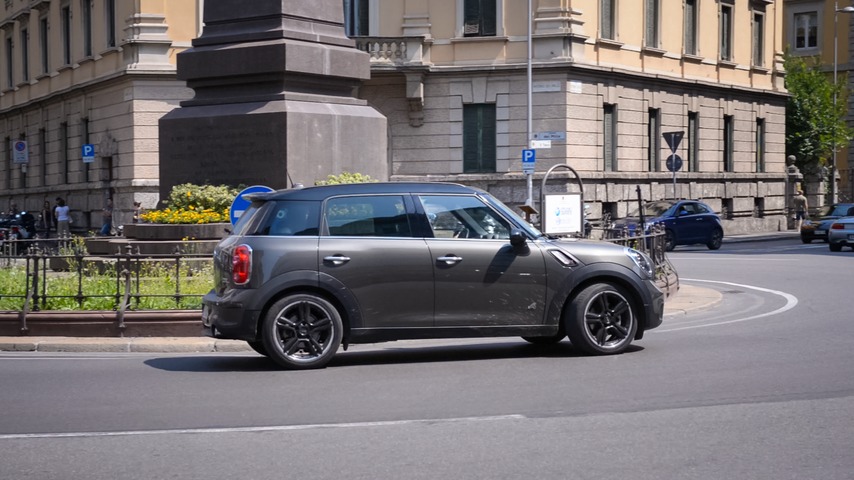} &
   \includegraphics[width = .33\linewidth,height=.2\linewidth]{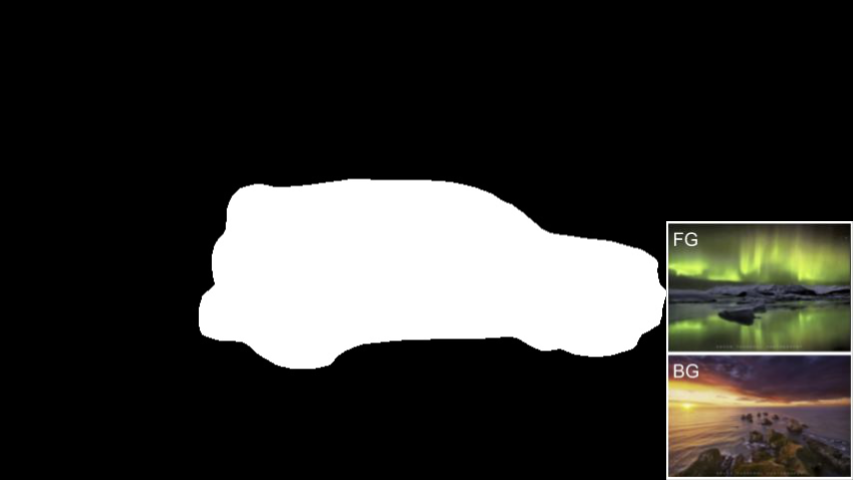} &
   \includegraphics[width = .33\linewidth,height=.2\linewidth]{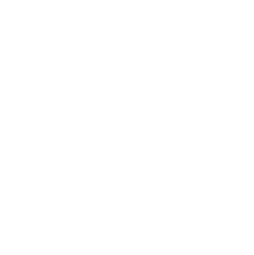} \\
  Content & Mask and styles &   \\
  \\
   \includegraphics[width = .33\linewidth,height=.2\linewidth]{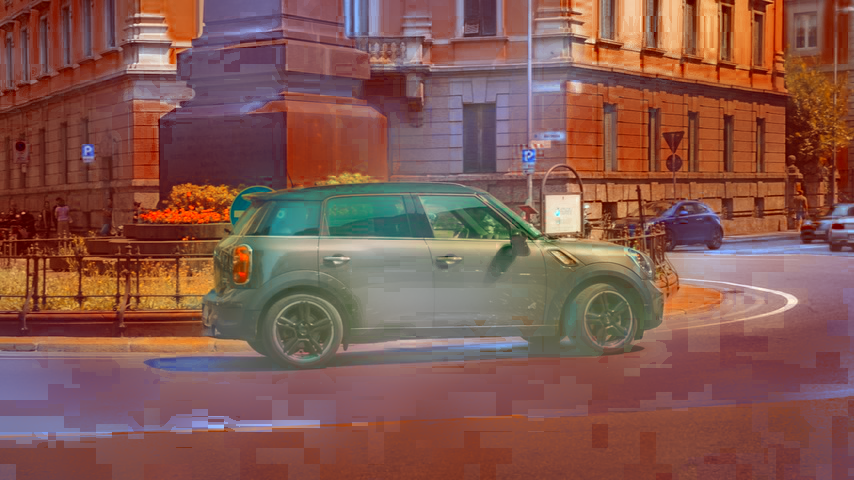} &
   \includegraphics[width = .33\linewidth,height=.2\linewidth]{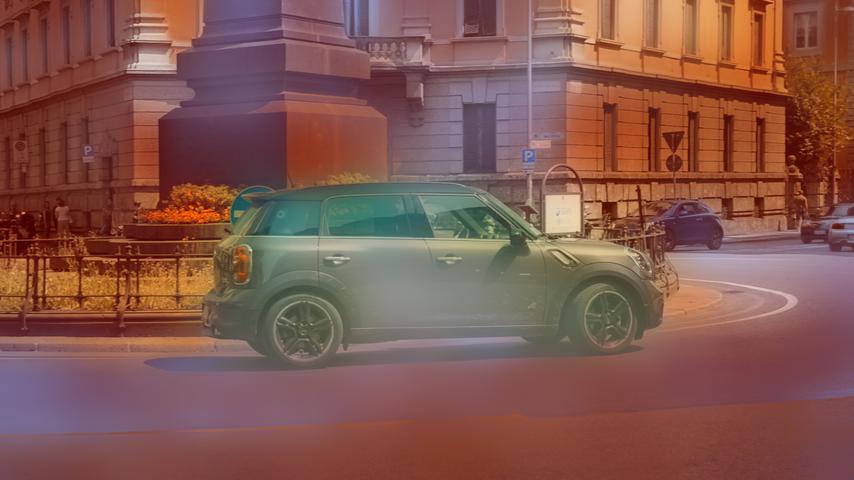} &
   \includegraphics[width = .33\linewidth,height=.2\linewidth]{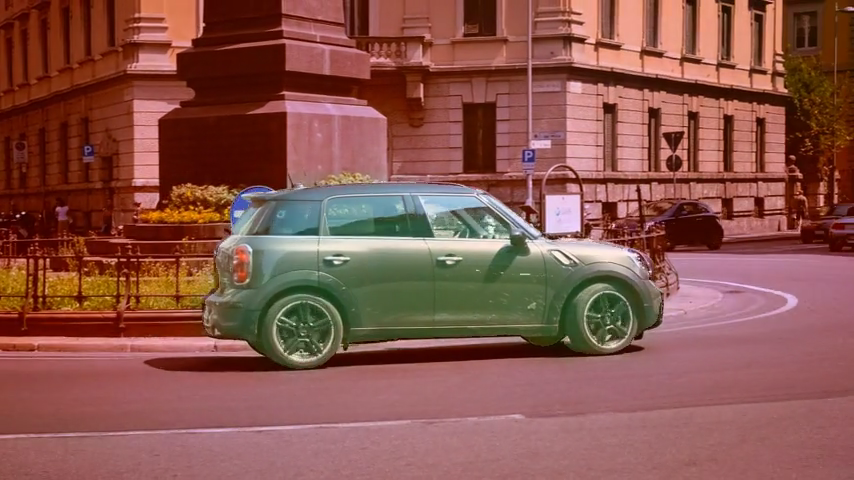} \\
  PhotoWCT~\cite{li2018closed} w/o spatial smoothing &PhotoWCT~\cite{li2018closed} w/ spatial smoothing &  Ours \\

\end{tabular}

\caption{Visual comparison with PhotoWCT~\cite{li2018closed}.}
\label{fig:photowct_smooth}
\vspace{1cm}
\end{figure*}

\begin{figure*}[ht!]
\centering
\begin{tabular}{c@{\hspace{0.005\linewidth}}c@{\hspace{0.005\linewidth}}c@{\hspace{0.005\linewidth}}c}

  \includegraphics[width = .33\linewidth,height=.2\linewidth]{images/supp_baseline_smooth/input_cows.jpg} &
   \includegraphics[width = .33\linewidth,height=.2\linewidth]{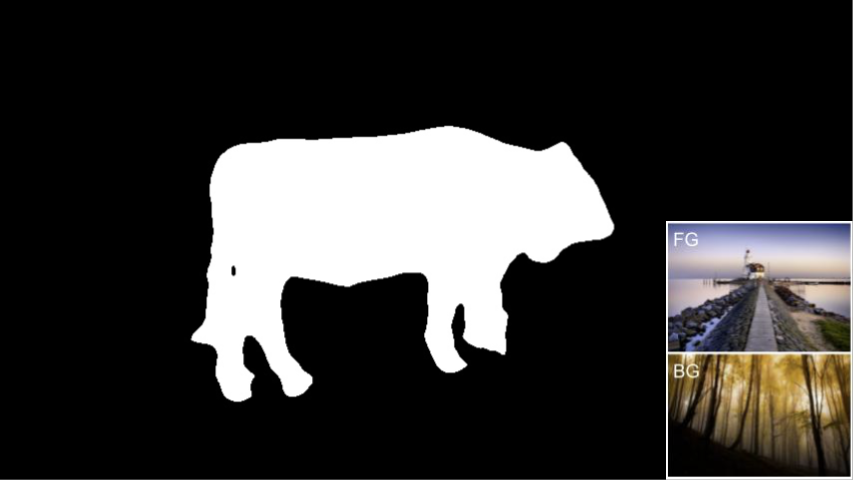} &
   \includegraphics[width = .33\linewidth,height=.2\linewidth]{images/supp_baseline_smooth/empty.jpg} \\
  Content & Mask and styles &   \\
  \\
   \includegraphics[width = .33\linewidth,height=.2\linewidth]{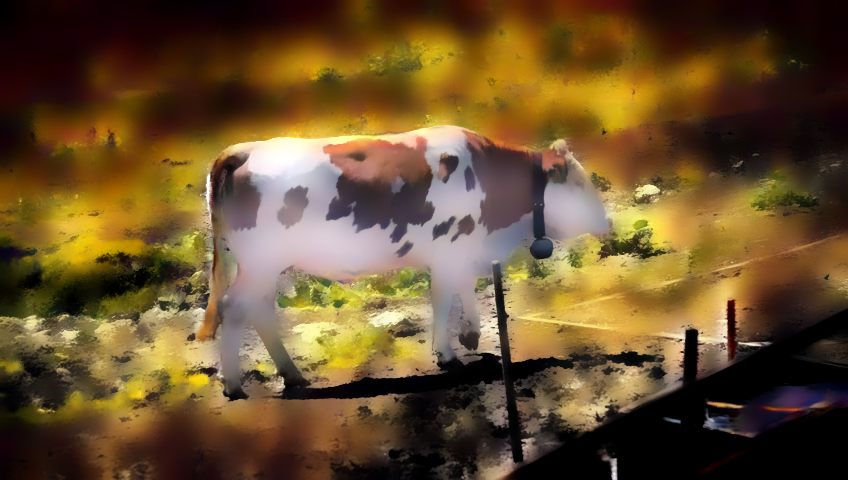} &
   \includegraphics[width = .33\linewidth,height=.2\linewidth]{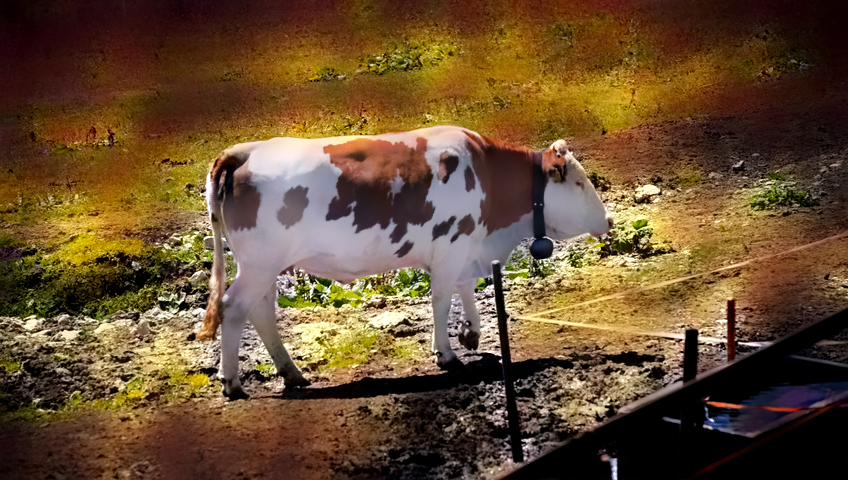} &
   \includegraphics[width = .33\linewidth,height=.2\linewidth]{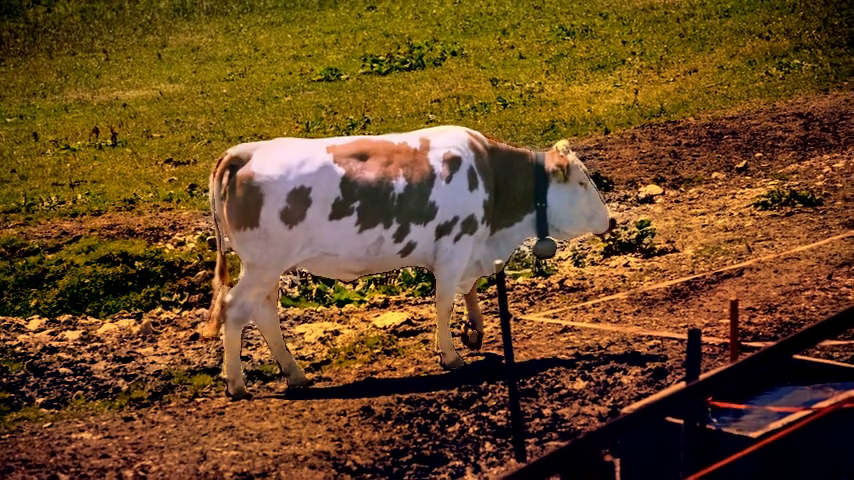} \\
  LST~\cite{li2019learning} w/o SPN & LST~\cite{li2019learning} w/ SPN &  Ours \\

\end{tabular}

\caption{Visual comparison with LST~\cite{li2019learning}.}
\label{fig:lst_spn}
\vspace{1cm}
\end{figure*}

\begin{figure*}[ht!]
\centering
\begin{tabular}{c@{\hspace{0.005\linewidth}}c@{\hspace{0.005\linewidth}}c@{\hspace{0.005\linewidth}}c}
  
  \includegraphics[width = .33\linewidth,height=.2\linewidth]{images/supp_baseline_smooth/input_swan.jpg} &
   \includegraphics[width = .33\linewidth,height=.2\linewidth]{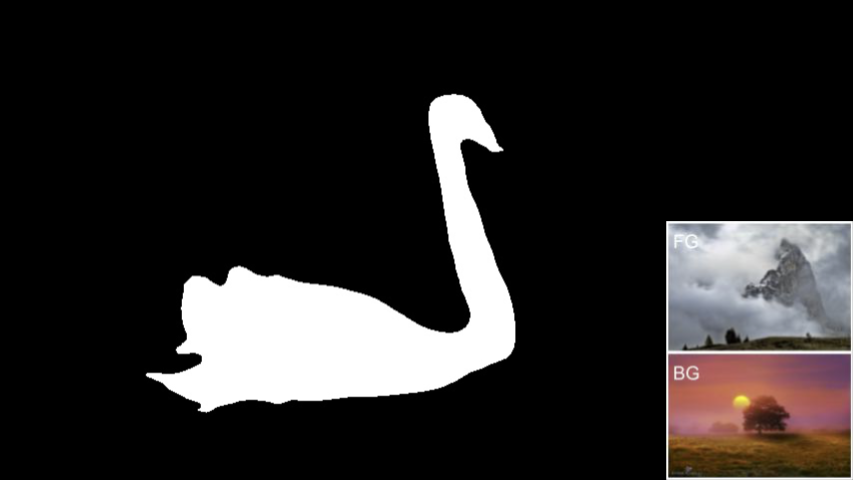} &
   \includegraphics[width = .33\linewidth,height=.2\linewidth]{images/supp_baseline_smooth/empty.jpg} \\
  Content & Mask and styles &   \\
  \\
  
   \includegraphics[width = .33\linewidth,height=.2\linewidth]{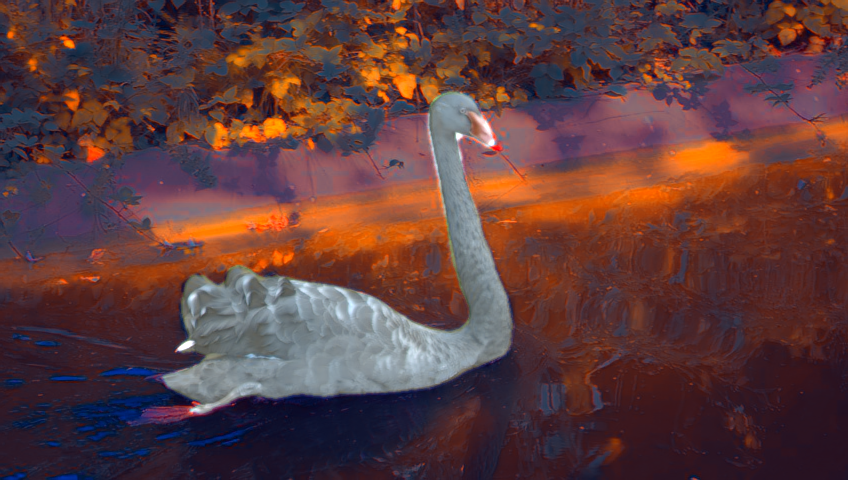} &
   \includegraphics[width = .33\linewidth,height=.2\linewidth]{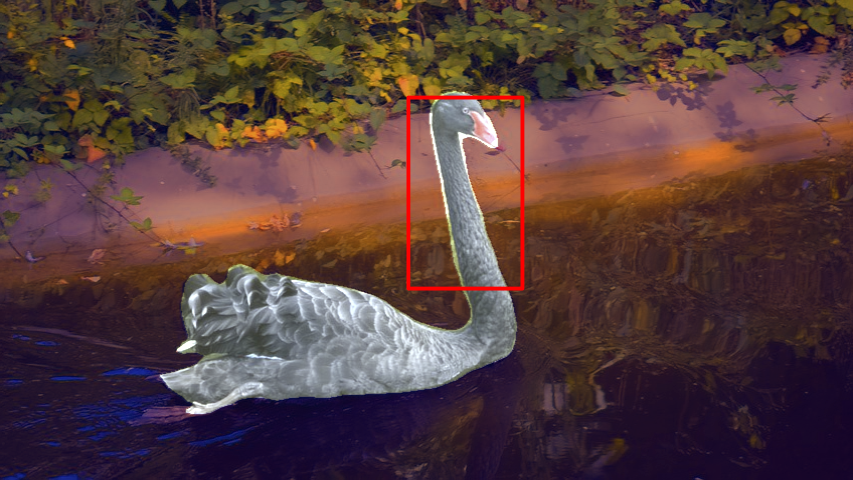} &
   \includegraphics[width = .33\linewidth,height=.2\linewidth]{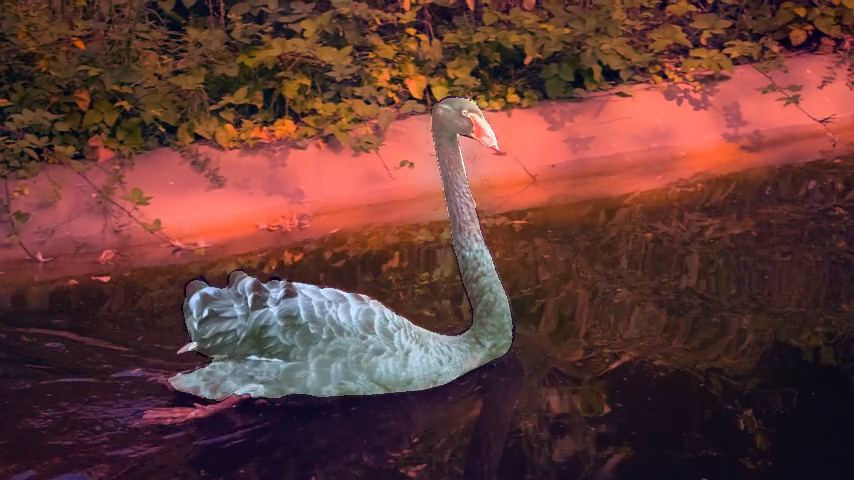}\\
  WCT\textsuperscript{2} w/o high-freq skips & WCT\textsuperscript{2} w/ high-freq skips &  Ours \\
   
\end{tabular}

\caption{Visual comparison with WCT\textsuperscript{2}.}
\label{fig:wct2_skip}
\end{figure*}

Here, we show visual comparisons to existing methods with and without applying their anti-distortion modules.
%
%
Figure~\ref{fig:photowct_smooth} shows the comparison with PhotoWCT.
Although the smoothing step helps reduce local artifacts, the smoothed result becomes blurry and hazy.
In contrast, our result preserves all the edges and image structures well.
In Figure~\ref{fig:lst_spn}, the result of LST without applying SPN has severe spatial distortions.
The SPN recovers some image details but cannot suppress all the distortions, resulting in an unrealistic result.
Figure~\ref{fig:wct2_skip} shows the results of WCT\textsuperscript{2} with and without the skip connections for high-frequency components.
It is clear that the skip connections helps bring back the image details and preserve the photorealism, showing comparable results to the proposed method.
More video comparisons are provided in the supplementary videos $visual\_comparison.mp4$ in the $\textbf{video\_mp4}$ folder.



\subsection{High-resolution style transfer}
\begin{figure*}[ht!]
\centering
\begin{tabular}{c@{\hspace{0.005\linewidth}}c@{\hspace{0.005\linewidth}}c@{\hspace{0.005\linewidth}}c}
  
  \includegraphics[width = .33\linewidth,height=.2\linewidth]{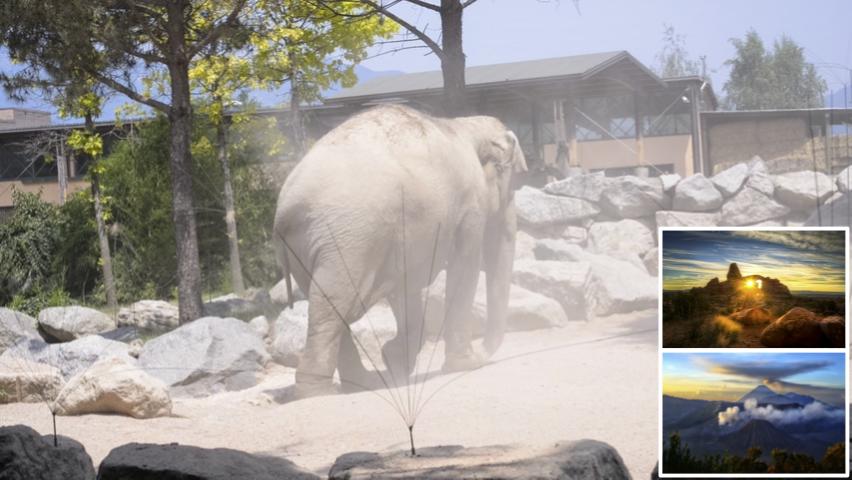} &
   \includegraphics[width = .33\linewidth,height=.2\linewidth]{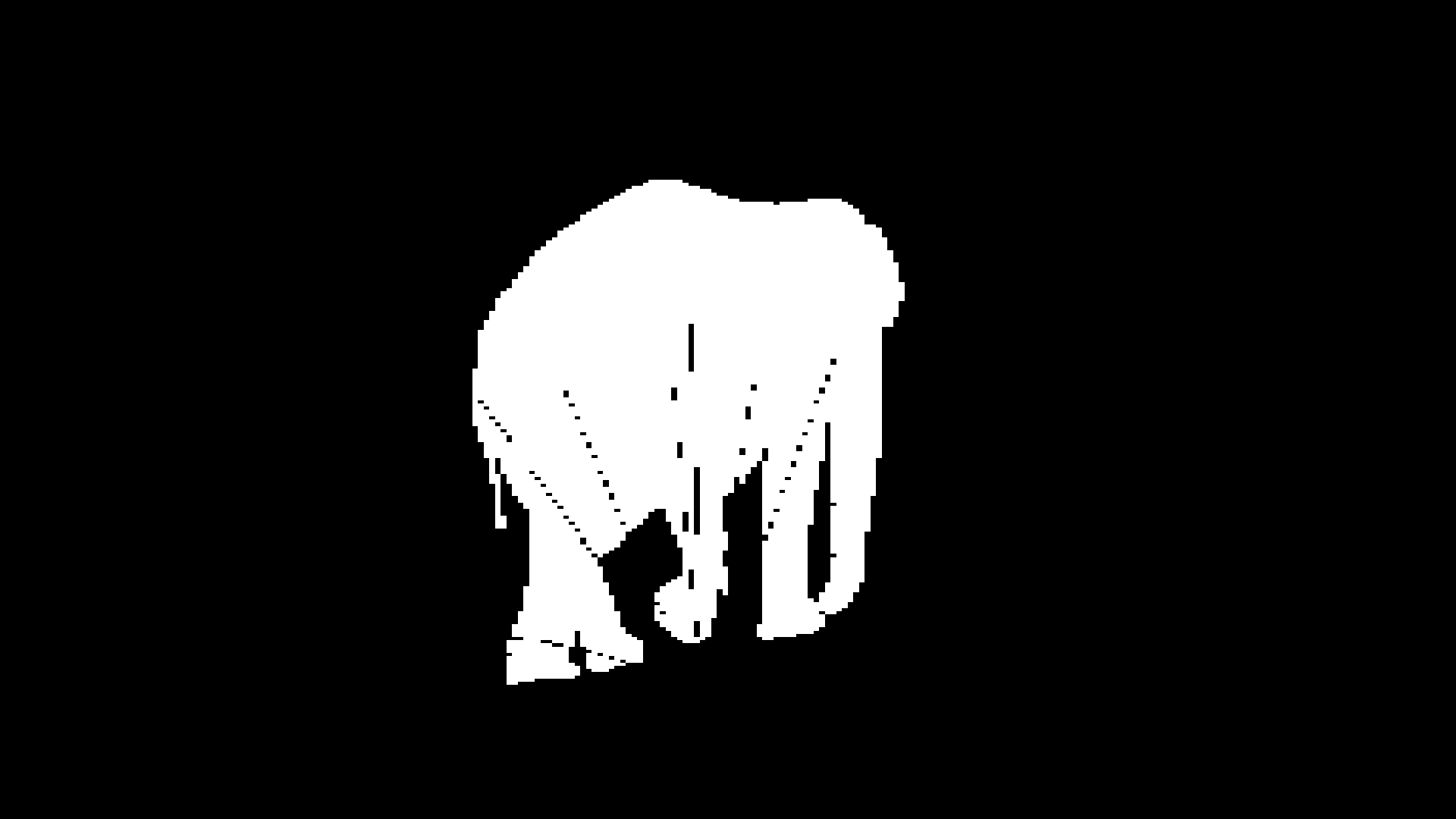} &
   \includegraphics[width = .33\linewidth,height=.2\linewidth]{images/supp_baseline_smooth/empty.jpg} \\
  Content and styles & Up-scaled mask  &   \\
  \\
  
   \includegraphics[width = .33\linewidth,height=.2\linewidth]{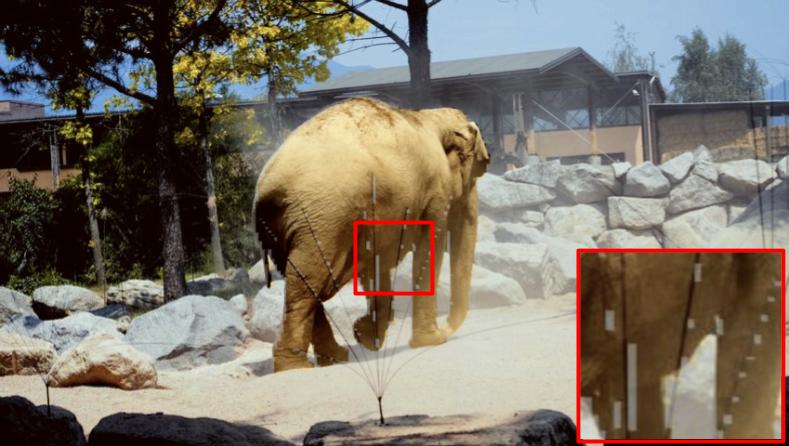} &
   \includegraphics[width = .33\linewidth,height=.2\linewidth]{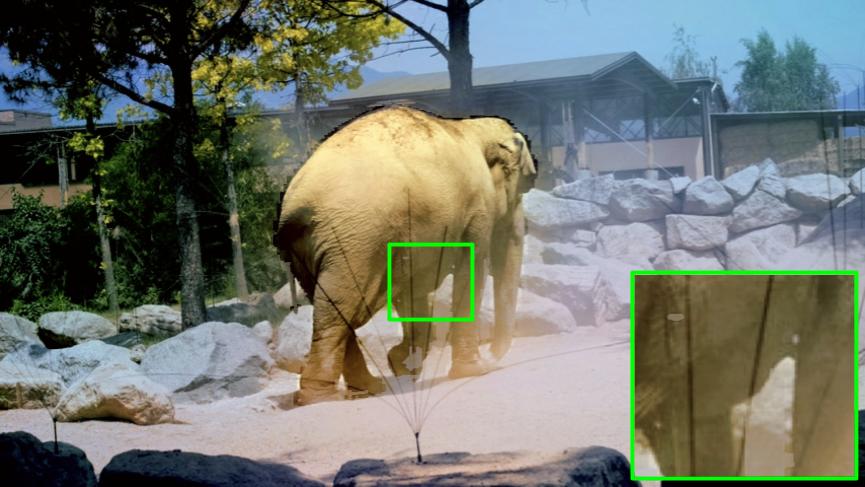} &
   \includegraphics[width = .33\linewidth,height=.2\linewidth]{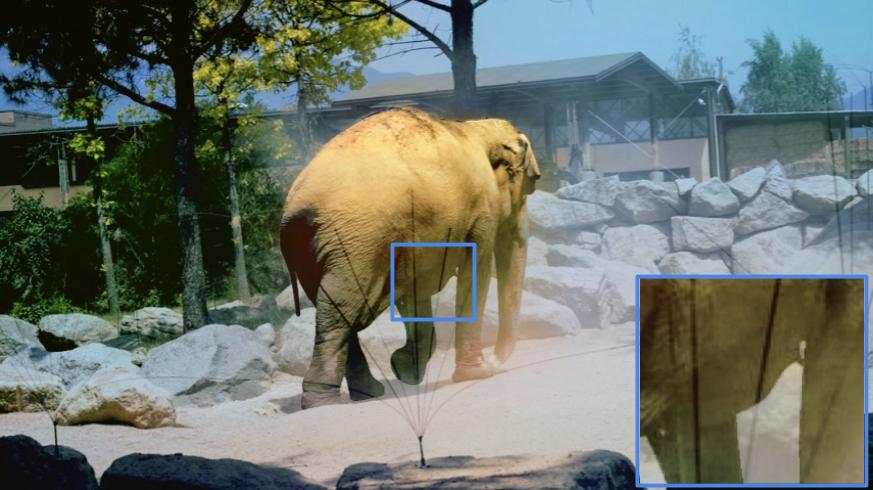} \\
  WCT\textsuperscript{2} with up-scaled mask  & Ours with upscaled mask  &  Ours with GT mask\\
   \\

   
\end{tabular}

\caption{High-resolution video stylization using low-resolution masks.}
\label{fig:mask_upscale}
\end{figure*}

Although the proposed method is efficient when processing high-resolution (e.g., $2000 \times 2000$) videos, extracting the object segmentation masks could be computationally expensive.
For example, it takes 2.61 seconds for a segmentation model, DeepLab~\cite{deeplabv3}, to process a single $1920 \times 1080$ frame.
Therefore, we aim to understand the feasibility of applying low-resolution segmentation masks for high-resolution style transfer.
First, we compute the object masks from a low-resolution ($256 \times 256$) input frame, which takes only 0.47 seconds for DeepLab.
Then, we resize the masks to $1920 \times 1080$ through nearest neighbor interpolation for transferring styles.
We compare the proposed method with WCT\textsuperscript{2} in Figure~\ref{fig:mask_upscale}.
The results from WCT\textsuperscript{2} have clear visual artifacts on the object boundaries as the upscaled masks are noisy.
On the other hand, our method is able to generate high-quality results, which are comparable to the ones produced by using the ground-truth object masks.
We demonstrates that our model performs well on high-resolution videos even if the segmentation masks is extracted from a low-resolution space. 
Please refer to the supplementary videos $HD\_video\_w\_upscaled\_mask\_*.mp4$ in the $\textbf{video\_mp4}$ folder for video comparisons.


\begin{figure*}[ht]
\centering
\begin{tabular}{c@{\hspace{0.005\linewidth}}c@{\hspace{0.005\linewidth}}c}
  
  \includegraphics[width = .33\linewidth,height=.18\linewidth]{images/failure_case/park_jpeg_00000.jpg} &
  \includegraphics[width = .33\linewidth,height=.18\linewidth]{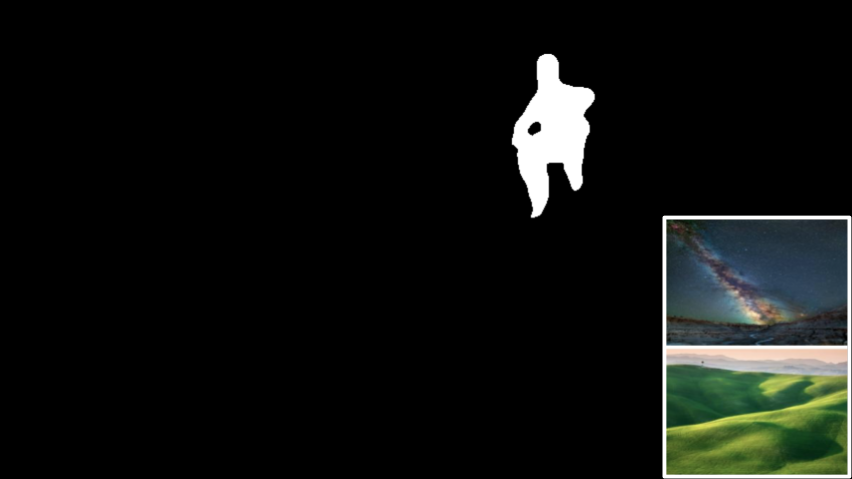} &
  \includegraphics[width = .33\linewidth,height=.18\linewidth]{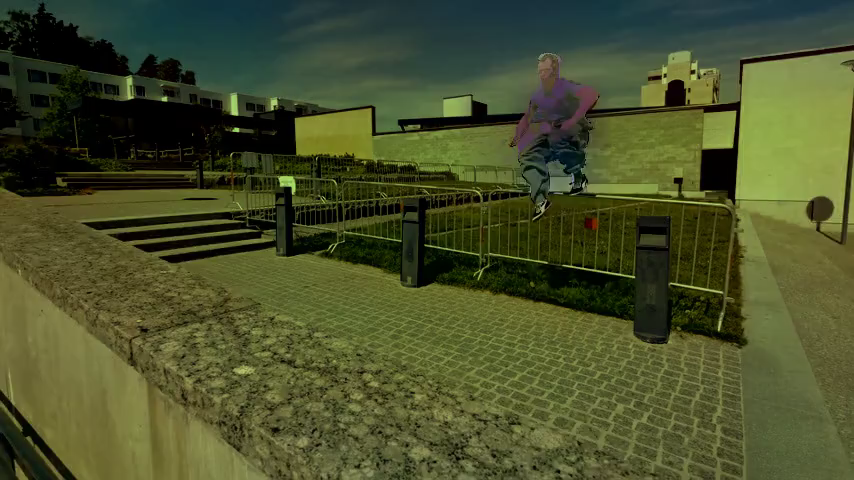}  \\
 
  \includegraphics[width = .33\linewidth,height=.18\linewidth]{images/failure_case/00057.jpg} &
  \includegraphics[width = .33\linewidth,height=.18\linewidth]{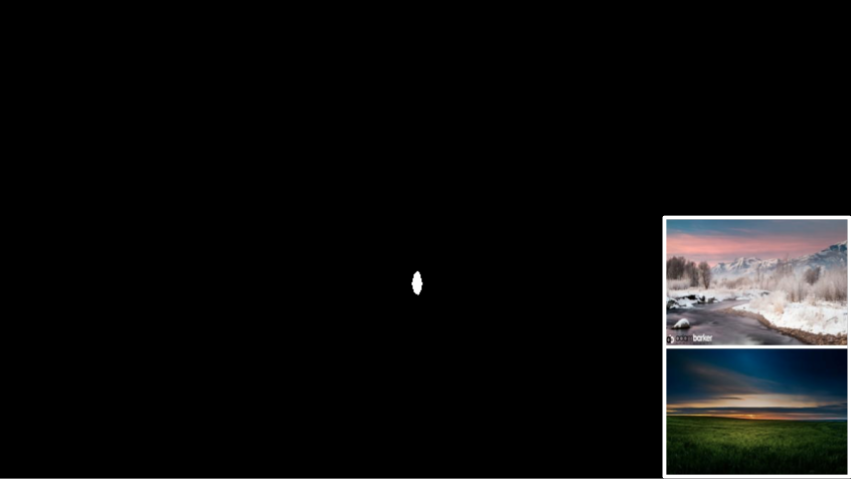} &
  \includegraphics[width = .33\linewidth,height=.18\linewidth]{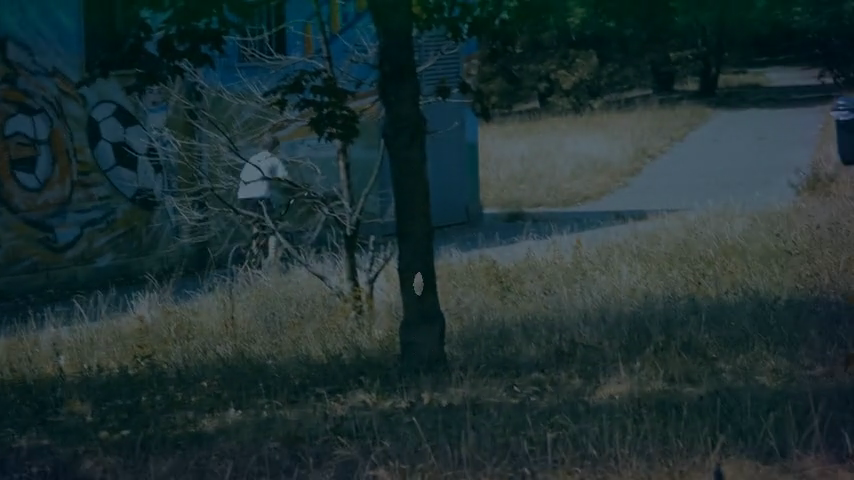}  \\

  Content  &  Mask and styles & Our result \\
  
\end{tabular}
\caption{Failure cases.}
\label{fig:fail_case}
\end{figure*}

\section{Failure Cases}
Our method requires reliable image-space statistics and can cope with modest amounts of segmentation noise. However, it may fail when the selected areas are too small (e.g., a textureless region with a single color).
On the other hand, when the foreground objects cannot be detected properly due to occlusion or reflection, our method may not be able to separate the styles of foreground and background very well, as shown in Figure~\ref{fig:fail_case}.


\clearpage
\clearpage

{\small
\bibliographystyle{ieee_fullname}
\bibliography{egbib}
}

\end{document}